%% file: main.tex
\pdfoutput=1
\def\year{2021}\relax

\documentclass[letterpaper]{article} 
\usepackage{aaai21}  
\usepackage{times}  
\usepackage{helvet}
\usepackage{courier}
\usepackage[hyphens]{url}
\usepackage{graphicx}
\usepackage{natbib}
\usepackage{caption}
\usepackage[switch]{lineno} 

\usepackage{hyperref}
\usepackage{eso-pic}
\usepackage{xspace}

\newcommand{\ouralgfull}{Semantic MapNet\xspace}
\newcommand{\ouralg}{SMNet\xspace}

\newcommand{\egoencoder}{Egocentric Visual Encoder\xspace}
\newcommand{\projector}{Feature Projector\xspace}
\newcommand{\memory}{Spatial Memory Tensor\xspace}
\newcommand{\decoder}{Map Decoder\xspace}

\newcommand{\beq}{\begin{eqnarray*}}
\newcommand{\eeq}{\end{eqnarray*}}
\newcommand{\beqn}{\begin{eqnarray}}
\newcommand{\eeqn}{\end{eqnarray}}
\newcommand{\ben}{\begin{enumerate}}
\newcommand{\een}{\end{enumerate}}
\newcommand{\bit}{\begin{itemize}}
\newcommand{\eit}{\end{itemize}}

\usepackage{eso-pic}
\usepackage{xspace}
\makeatletter
\DeclareRobustCommand\onedot{\futurelet\@let@token\@onedot}
\def\@onedot{\ifx\@let@token.\else.\null\fi\xspace}

\def\eg{\emph{e.g}\onedot} 
\def\ie{\emph{i.e}\onedot} 

\def\etc{\emph{etc}\onedot} \def\vs{\emph{vs}\onedot}
\def\wrt{w.r.t\onedot} 
\def\etal{\emph{et al}\onedot}
\makeatother

\newcommand{\figref}[1]{Fig.~\ref{#1}}
\newcommand{\eqnref}[1]{Eqn.~\ref{#1}}

\newcommand{\tableref}[1]{Table \ref{#1}}

\newcommand{\myquote}[1]{\emph{`#1'}}
\newcommand{\myapprox}{{\raise.17ex\hbox{$\scriptstyle\sim$}}}

\newcommand{\xhdr}[1]{\vspace{1.75pt} \noindent {\textbf{#1}}}

\makeatletter
\newcommand{\bfgreek}[1]{\bm{\@nameuse{#1}}}
\makeatother

\usepackage{comment}
\usepackage{amsmath,amssymb} 
\usepackage{color}
\usepackage{tikz}
\usepackage[belowskip=0pt,aboveskip=0pt,font=small]{caption}
\usepackage[belowskip=0pt,aboveskip=0pt,font=small]{subcaption}
\usepackage{makecell}
\usepackage{color,soul}

\usepackage{multirow}
\usepackage{rotating}
\usepackage{booktabs}
\usepackage{tabularx}
\usepackage{xcolor}
\usepackage{colortbl}
\newcolumntype{s}{>{\columncolor[gray]{.85}[.5\tabcolsep]}c}
\usepackage[colorinlistoftodos]{todonotes}

\definecolor{void}{RGB}{0,0,0}
\definecolor{shelving}{RGB}{106, 137, 204}
\definecolor{chestofdrawers}{RGB}{230, 126, 34}
\definecolor{bed}{RGB}{7, 153, 146}
\definecolor{cushion}{RGB}{248, 194, 145}
\definecolor{fireplace}{RGB}{76, 209, 55}
\definecolor{sofa}{RGB}{255, 168, 1}
\definecolor{table}{RGB}{184, 233, 148}
\definecolor{chair}{RGB}{39, 174, 96}
\definecolor{cabinet}{RGB}{229, 80, 57}
\definecolor{plant}{RGB}{30, 55, 153}
\definecolor{counter}{RGB}{24, 220, 255}
\definecolor{sink}{RGB}{234, 32, 39}

\usepackage{paralist}

\title{\ouralgfull: Building Allocentric Semantic \\Maps and Representations from Egocentric Views}
\author {
    Vincent Cartillier,\textsuperscript{\rm 1}
    Zhile Ren,\textsuperscript{\rm 1}
    Neha Jain,\textsuperscript{\rm 1}
    Stefan Lee,\textsuperscript{\rm 2}
    Irfan Essa,\textsuperscript{\rm 1,4}
    Dhruv Batra,\textsuperscript{\rm 1,3} \\
}
\affiliations {
    \textsuperscript{\rm 1} Georgia Institute of Technology \\     
    \textsuperscript{\rm 2} Oregon State University \\ 
    \textsuperscript{\rm 3} Facebook AI Research \\
    \textsuperscript{\rm 4} Google Research \\
    
    \{vcartillier3, zren80, nehaj, irfan, dbatra\}@gatech.edu, leestef@oregonstate.edu
}

\begin{document}
\maketitle

\begin{abstract}


We study the task of \emph{semantic mapping} -- specifically, an embodied agent (a robot or an egocentric AI assistant) 
is given a tour of a new environment and asked to build an allocentric top-down semantic map (\myquote{what is where?}) from egocentric observations of an RGB-D camera with known pose (via localization sensors). 
Towards this goal, we present \ouralgfull (\ouralg), which 
consists of: 
\begin{inparaenum}[(1)]
\item an \egoencoder that encodes each egocentric RGB-D frame, 
\item a \projector that 
projects egocentric features to appropriate locations on a floor-plan, 
\item a \memory
of size floor-plan length $\times$ width $\times$ feature-dims that learns to accumulate projected egocentric features, and
\item a \decoder that uses the memory tensor to produce semantic top-down maps. 
\end{inparaenum} 
\ouralg combines the strengths of (known) projective camera geometry and neural representation learning. On the task of semantic mapping in the Matterport3D dataset, \ouralg significantly outperforms competitive baselines by $4.01-16.81$\% (absolute) on mean-IoU and $3.81-19.69$\% (absolute) on Boundary-F1 metrics. 
Moreover, we show how to use the neural episodic memories and spatio-semantic allocentric representations build by \ouralg 
for subsequent tasks in the same space -- navigating to objects seen during the tour (\myquote{Find chair}) or answering questions about the space (\myquote{How many chairs did you see in the house?}). 
Project page: \href{https://vincentcartillier.github.io/smnet.html}{\textcolor{purple}{https://vincentcartillier.github.io/smnet.html}}.
\end{abstract}


\section{Introduction}

Imagine yourself receiving a tour of a new environment.  
Maybe you visit a friend's new house and they show you around (\myquote{This is our living room, and down here is the study}). 
Or maybe you accompany a real-estate agent as they show you a new office space (\myquote{These are the cubicles, and down here is the conference room}). Or someone gives you a tour of a mall or a commercial complex. In all these situations, humans have the ability to form episodic memories and spatio-semantic representations of these spaces \cite{o1978hippocampus}. 
We can recall which spaces we visited (living room, kitchen, bedroom, \etc), what objects were present (chairs, tables, whiteboards, \etc), and what their relative arrangements were (the kitchen was combined with the open plan living room, the bedroom was down the hallway, \etc). 
We can also leverage these representations to perform new tasks in these spaces (\eg navigate to the restroom via a path shorter than the one demonstrated on the tour). Of course, human memory is limited in time and in the level of metric detail it stores \cite{epstein_natureneuro17}. 
Our long-term goal 
is to develop super-human AI agents that can build 
rich, accurate, and reusable spatio-semantic representations from egocentric observations. This capability is an essential building block 
for autonomous navigation, mobile manipulation, and egocentric personal AI assistants. 

\begin{figure}[t]
    \centering
    \includegraphics[width=1\linewidth]{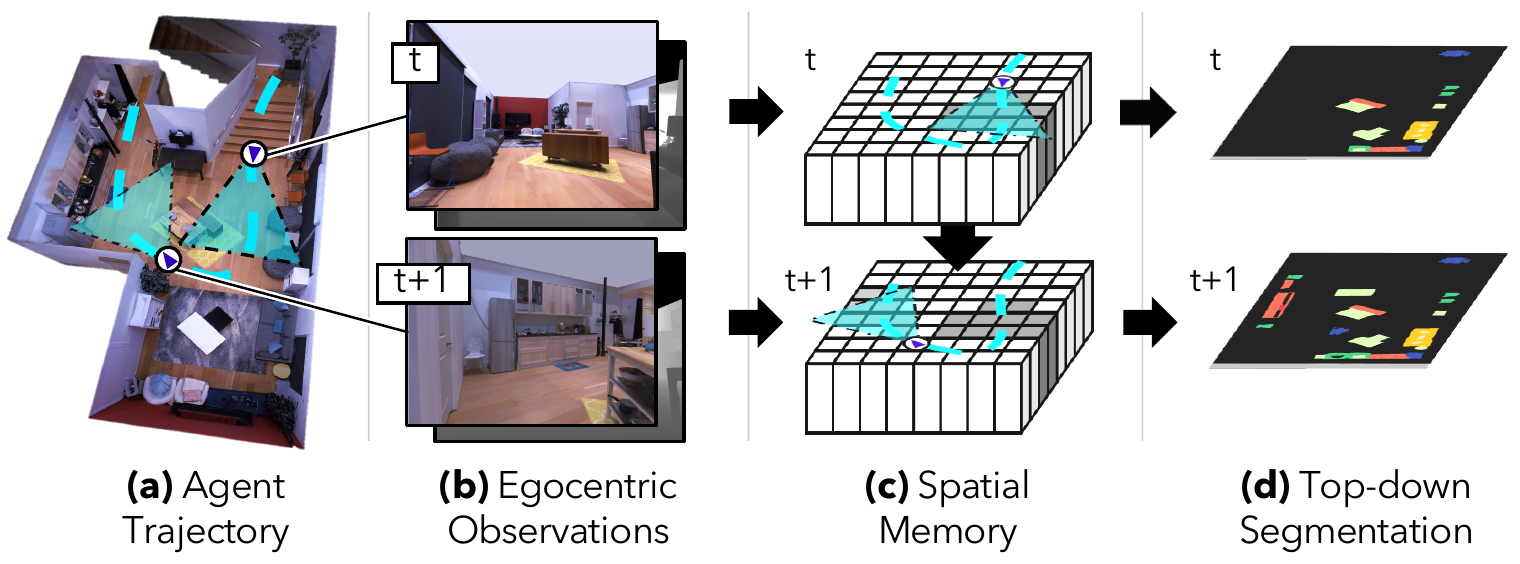}\\[5pt]
    \caption{Semantic Mapping: (a) While moving through a 3D space (with known pose), our agent converts egocentric RGB-D observations (b) to representations in an allocentric spatial memory (c), 
    which is used to predict top-down semantic segmentations (d) showing \myquote{what objects are where} from a birds-eye view. 
   } 
\label{fig:teaser}
\end{figure}

In this paper, we study the specific task of creating an allocentric top-down semantic map of an indoor space, illustrated in \figref{fig:teaser}.  
An embodied agent (a virtual robot or an egocentric AI assistant) is equipped with an RGB-D camera with known pose (extracted via localization sensors such as  
GPS and IMU). The agent is provided a tour of a new environment, represented as a trajectory of camera poses (shown in \figref{fig:teaser}(a)). 
The task then is to produce an allocentric top-down semantic map (shown in \figref{fig:teaser}(d)) from 
the sequence of egocentric observations with known pose (shown in \figref{fig:teaser}(b)). 
Our experiments focus on top-down \emph{semantic segmentation}, \ie each pixel in the top-down map is assigned to a single class label (of the \emph{tallest} object at that location on the floor, \ie the one visible from the top-down view). 
Our produced semantic maps are \emph{metric} --  
each pixel corresponds to a 2cm $\times$ 2cm grid on the floor -- as opposed to topological maps \cite{fraundorfer_icirs07, ego-topo} 
that lack spatial information (scale, position) and do not support our downstream tasks of interest. Importantly, while the semantic top-down map is our primary `product', our goal is to build neural episodic memories and spatio-semantic \emph{representations} of 3D spaces in the process. Representations that enable 
the agent to easily learn and accomplish subsequent tasks in the same space -- navigating to objects seen during the tour 
(\myquote{Go to a chair}) or answering questions about the space (\myquote{How many chairs did you see in the house?}). 

\begin{figure}[t]
    \centering
    \includegraphics[width=1\linewidth]{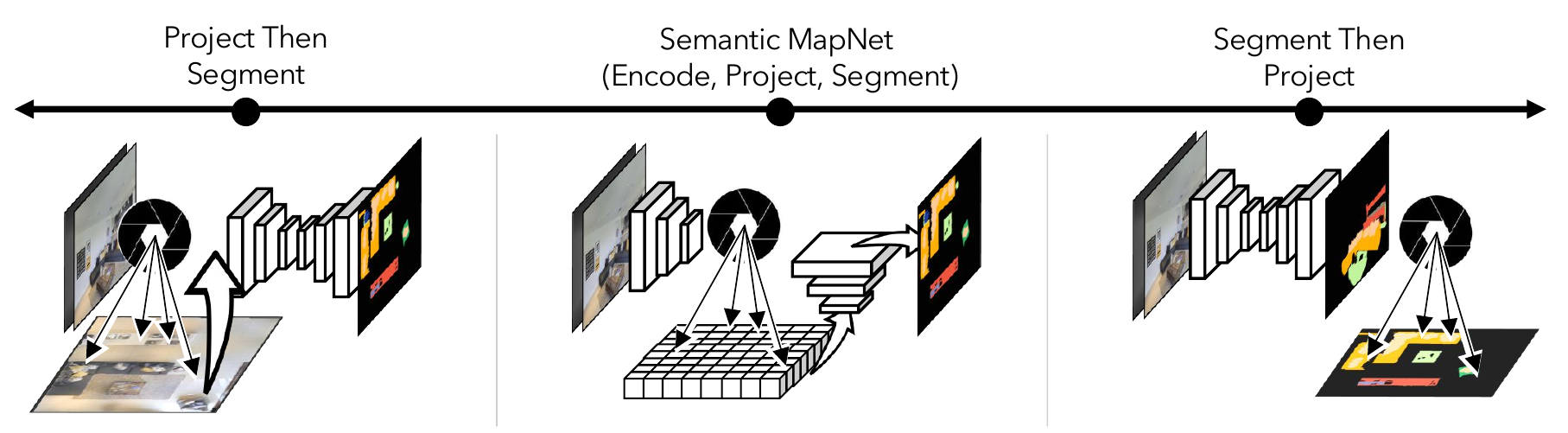}\\[2pt]
    \caption{A spectrum of approaches to top-down semantic segmentation: (Right) perform egocentric semantic segmentation and project \emph{labels}; (Left) Construct an overhead imagery (project \emph{pixels}) and perform semantic segmentation; (Middle) \ouralg: encode pixels, project \emph{features}, decode labels.} 
\label{fig:related}
\end{figure}

\textbf{What should we project?} 
Approaches to top-down or overhead semantic segmentation can be arranged on spectrum illustrated in \figref{fig:related}. At one end (on the right), are approaches \cite{sengupta_icirs12, sunderhauf_icra16, maturana_fsr18} that first perform egocentric semantic segmentation and then use the known camera pose and the depth of each pixel to project \emph{labels} to an allocentric map. 
In our experiments, we find that this results in `label splatter' -- any mistakes in the egocentric semantic segmentation made at the depth boundaries of objects get splattered on the map around the object. This problem can be slightly assuaged via image processing heuristics (median filtering). However, the fundamental issue 
persists even after those `bells and whistles'.Quantitatively, this results in high precision but low recall of the object segmentation. At the other end of the spectrum (on left) are approaches that operate on a single overhead image (projecting \emph{pixels} if needed) and perform semantic segmentation on this image  
\cite{singh_bmvc18, mattyus_iccv15}. 
While this may be appropriate for aerial or geospatial imagery, converting multiple high-res egocentric images 
into a single bird's eye view is wasteful and throws out significant visual information. Qualitatively, we find that this results in coarse segmentations; object sizes are under-estimated, small objects missed entirely. 
Quantitatively, we see high precision but low recall. 

We pursue an approach called \ouralgfull (\ouralg) that lies in the middle of this spectrum.
Specifically, as shown in \figref{fig:related} (middle), \ouralg extracts visual \emph{features} in the egocentric reference frame, but predicts semantic segmentation \emph{labels} in the allocentric reference frame. This is accomplished by projecting egocentric features to appropriate locations in an allocentric spatial memory, and using this memory to decode a top-down semantic segmentation.  
This design addresses both deficiencies in prior work -- (a) the spatial-memory-to-map decoder in \ouralg is based on transposed convolutions and learns to smooth out any `feature splatter'; and (b) the egocentric feature extractor in \ouralg operates directly on high-res egocentric images and is able to recognize and segment small objects that may not be visible from a bird's eye view. 

We conduct experiments on the photo-realistic scans of building-scale environments (homes, offices, churches) in the Matterport3D dataset~\cite{chang2017matterport3d} using the Habitat simulation platform~\cite{habitat19iccv} (giving us access to agent state, navigation trajectories, RGB-D renderings, \etc). 
We choose the Matterport3D dataset because it provides semantic annotations in 3D, the spaces are large enough to allow multi-room traversal by the agent, and the use of a 3D simulator allows us to render RGB-D from any viewpoint, create top-down semantic annotations, and study embodied AI applications in the same environments.
Quantitatively, on the task of semantic mapping, \ouralg significantly outperforms the aforementioned baselines
by $4.01-16.81$\% on mean-IoU and $3.81-19.69$\% (absolute) on Boundary-F1 metrics

\ouralg combines the strengths of (known) projective camera geometry with neural representation learning, and address our key desideratum -- learning rich, reusable spatio-semantic representations.
We demonstrate via extension experiments how representations built by \ouralg from a single tour of an environment can be reused for ObjectNav and Embodied Question Answering~\cite{das2018embodied}.


\section{Related Work}

\begin{figure*}[t]
    \centering
    \includegraphics[width=0.95\linewidth]{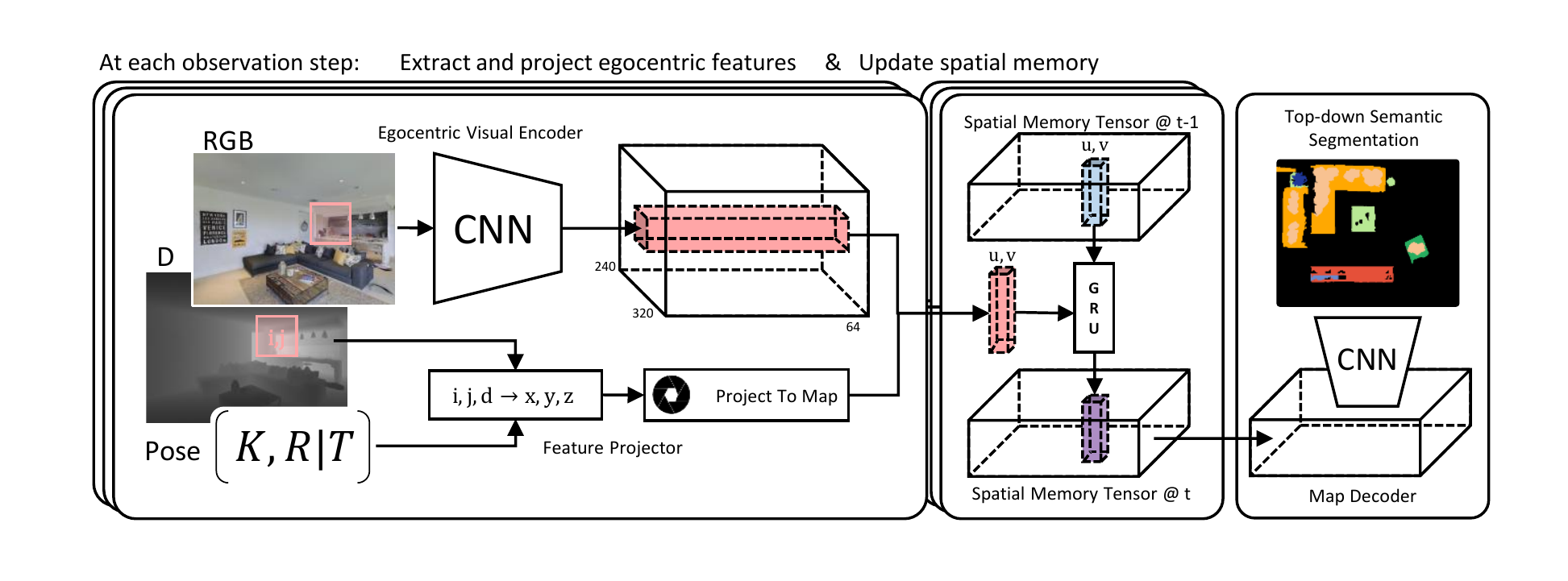}\\[-10pt]
   \caption{At each step in a trajectory, SMNet updates an allocentric map based on egocentric observations. Egocentric RGB-D observations are represented using a CNN encoder and the feature vectors are projected to the spatial memory (left). Memory cells are updated by a GRU to incorporate this new information (middle). The spatial memory can then be decoded by to perform top-down semantic segmentation (right).} 
\label{fig:model}
\end{figure*}

\xhdr{Spatial Episodic Memories for Embodied Agents.} Building and dynamically updating a spatial memory is a powerful inductive bias that has been studied in many embodied settings. Most SLAM systems perform localization by registration to sets of localized keypoint features \cite{mur2017orb}. Many recent works in embodied AI have developed agents for navigation \cite{anderson2019chasing,beeching2020egomap,gupta2017cognitive,georgakis2019simultaneous,blukis2018mapping} and localization \cite{henriques2018mapnet,parisotto2017neural,zhang2017neural} that build 2.5D spatial memories containing deep features from egocentric observation. Like our approach, these all involve some variation of egocentric feature extraction, pin-hole camera projection, and  map update mechanisms. However, these works focus on spatial memories as part of an end-to-end agent for a downstream task and do not evaluate the quality of the generated maps in terms of environment semantics directly, nor study how segmentation quality affects downstream tasks.

\xhdr{Semantic Mapping from Egocentric Observations.} Predicting top-down semantic segmentation from egocentric observations has been studied in the context of robotics as the semantic SLAM (or semantic mapping) problem \cite{rosinol2019kimera,maturana2018real,grinvald2019volumetric,mccormac2017semanticfusion}. We compare with a recent representative algorithm in this family as our baseline \cite{grinvald2019volumetric}. Further work has examined the use of semantic labels as an intermediate step in an end-to-end model \cite{gordon2018iqa,chaplot2020object} or to derive supervision to reward agent trajectories \cite{chaplot2020semantic}. These works have not evaluated the quality of the semantic map and instead focused on downstream tasks. All these works follow the Segment-then-Project paradigm -- invoking a segmentation network on the 2D observations and then projecting labels into an allocentric map. In contrast, our findings suggest it is more effective to project intermediate features and allow an allocentric decoder to produce the final segmentation.

Closely related to our approach is a line of work focusing on volumetric recurrent memory architectures for 3D semantic segmentation of small objects \cite{cheng2018geometry,tung2019learning}. Like \ouralgfull, these approaches project intermediate features into a spatial memory and then decode segmentations from that structure. However, these works focus on relatively small objects due to the large memory constraints of 3D volumetric memory. For example, a 25m $\times$ 20m footprint indoor environment with standard ceiling height (2.75m) would require storing 171.875 million feature vectors at 2cm$^3$ resolution -- or a total of 176 gigabytes if features are 256 dimensional as in our experiments.
\ouralgfull is designed to work on the large environments (average footprint of 24.5m $\times$ 23.4m) of the Matterport3D dataset, and achieves state of the art results.

Recently, \citeauthor{pan2020cross} examined cross-view semantic segmentation -- i.e. the task of predicting a local top-down semantic map from a single first-person observation \cite{pan2020cross}. Unlike SMNet, their proposed approach does not include projective geometry -- instead learning a small network to transform first-person views to top-down feature maps -- nor does it accumulate observations over a trajectory. In contrast, we addresses the problem of building a global top-down semantic map based on a trajectory.

\xhdr{Simulation Platforms and Embodied Vision Tasks.} The creation of large 3D datasets~\cite{chang2017matterport3d,armeni20163d} and simulators~\cite{habitat19iccv,mattersim,kolve2017ai2,xiazamirhe2018gibsonenv} has spurred development in Embodied AI. 
Recent work examines interactive agents navigating in the environment to answer questions~\cite{das2018embodied,wijmans2019embodied,gordon2018iqa}, reach desired locations~\cite{wijmans2019decentralized}, and infer the shape of occluded objects~\cite{yang2019embodied}. 
We study a fundamental building block for these tasks -- building top-down semantic maps of indoor environments.

\section{\ouralgfull (\ouralg)}
\label{sec:memory}

We now describe our proposed approach for semantic mapping, called \ouralgfull (\ouralg), in detail. 
As shown in \figref{fig:model}, \ouralg consists of the following modules: 
\begin{compactenum}[--]
\item An \textbf{\egoencoder} that converts each egocentric RGB-D frame into a $\mathbb{R}^{w \times h \times d}$ feature tensor, representing the content of each image region. 
\item A \textbf{\projector} that uses the known camera pose and the depth of each pixel to project these egocentric features at appropriate locations on a floor-plan,  
\item A \textbf{\memory} of size floor-plan length $\times$ width $\times$ feature-dims that accumulates these projected egocentric features. Repeated observations of the same spatial locations are incorporated through a learned recurrent model operating at each location. 
\item A \textbf{\decoder} that uses the accumulated memory tensor to produce top-down semantic segmentations. 
\end{compactenum} 

\xhdr{Problem Setup and Notation.} 
Let $I$ denote an RGB-D image. We assume a known camera -- specifically, let $K$ be the camera intrinsic matrix, and $[R \mid \mathbf{t}]$ denote the camera extrinsic matrix (rotation and translation needed to convert world coordinates to camera coordinates). 
Thus, an agent's trajectory through an environment is represented as a sequence of egocentric RGB-D observations $I^{(1)}, \dots, I^{(T)}$ at known poses $[R \mid \mathbf{t}]^{(1)}, \dots, [R \mid \mathbf{t}]^{(T)}$. Strictly speaking, our approach does not require knowing camera pose in world coordinates at all times -- all we need are successive pose transformations $[R \mid \mathbf{t}]^{(t\rightarrow t+1)}$, a problem known in robotics and computer vision as egomotion estimation. The entire approach could be defined in terms of the camera coordinates at time $t=1$. However, for sake of clarity of the exposition, we describe our approach with global pose. 

Let $S$ denote the top-down semantic segmentation. 
Each pixel in $S$ represents a 2cm$\times$2cm cell in the environment and is labeled with the class of the tallest object in that cell (\ie the object visible from above). 
At each time $t$, let $M^{(t)}$ denote the memory tensor incorporating all the information observed in the trajectory so far, and let $\hat{S}^{(t)}$ denote the segmentation predicted using $M^{(t)}$. 
Note that test-time evaluation is done using $\hat{S}^{(T)}$, but during training our agent predicts and receives supervision for intermediate predictions along the trajectory $\hat{S}^{(1)}, \ldots, \hat{S}^{(T)}$. 

There are a number of coordinate systems in this discussion which we define now for clarity -- pixel positions in the egocentric RGB-D image $I$ are indexed with $i,j$, and the depth at this pixel is denoted with $d_{i,j}$ (or $d$ when its clear from context which pixel is being talked about). 
A 3D point in world coordinates is denoted with $x,y,z$. 
For notational simplicity, we follow the standard convention in computer graphics -- negative $Y$-axis aligned gravity in the world coordinate system. 
Finally, cells in the memory tensor are indexed with $u,v$.  
%
Next, we describe each module in detail.

\xhdr{\egoencoder.}
Each egocentric frame $I^{(t)}$ gives a local glimpse of the environment -- providing information about objects and their locations in the current view. 
To represent these, we encode each RGB-D frame using RedNet~\cite{jiang2018rednet}, 
a recently proposed architecture for semantic segmentation of indoor scenes. 
In principle, one may choose any standard image encoder network for semantic segmentation such as Mask-RCNN~\cite{he2017mask}. We chose RedNet simply because the network structure has proven to be effective for parsing indoor environments and pre-trained models (learned on SUN-RGBD dataset~\cite{song2015sun}) are publicly available. 
We initialize with these pre-trained weights and fine-tune RedNet on our dataset. 
We conducted several experiments by extracting egocentric features at different stages in the RedNet network (encoder, last layer, scores, softmax, one-hot encoded labels). We found that encoding each RGB-D frame with the last layer RedNet features yields to the best performances.
The output of this encoder for image $I^{(t)}$ is an egocentric feature map $F^{(t)} \in \mathbb{R}^{240\times320\times64}$ with each of the $240\times320$ cells storing a 64-d feature. We upscale this tensor to the resolution of the depth image (480$\times$640) with bilinear interpolation, resulting in each pixel $i,j$ having an associated feature $F^{(t)}_{i,j} \in \mathbb{R}^{64}$ and depth value $d_{i,j}$.

\xhdr{\projector.}
To project an egocentric feature $F^{(t)}_{i,j}$ to the spatial memory, we must (a) shoot a ray from the camera center through the image pixel $(i,j)$ out to a depth $d_{i,j}$ to get a 3D point in the camera coordinate system, (b) convert from camera to world coordinates to get the corresponding $(x,y,z)$, 
and then (c) project it to cell indices $u,v$ in the memory tensor. With known camera pose and intrinsics, these transformations for the standard pinhole camera 
can be written compactly as: 
\begin{equation}
\underbrace{
\begin{bmatrix}
x\\
y\\
z
\end{bmatrix} =
d_{i,j} R^{-1} K^{-1}
\begin{bmatrix}
i\\
j\\
1
\end{bmatrix}
 - \mathbf{t},
}_{\text{(Inverse) Pinhole Camera Projection}} 
 \quad \text{and} \quad
\underbrace{
\begin{bmatrix}
u \\
v \\
0 \\
1
\end{bmatrix}
= P_v
\begin{bmatrix}
x\\
y\\
z\\
1
\end{bmatrix}
}_{\text{Orthographic Projection}} 
\label{eq:mapper}
\end{equation}
%
%
%
where $P_v$ is a known orthographic projection matrix converting 3D world coordinates to 2D memory cell indices. 
When several points are projected to the same index in $M^{(t)}$, we retain the one with the maximum height in the world coordinates. 
This results in a set of projected features $F^{(t)}_{u,v}$. 

\xhdr{\memory}
$M$ is a 3D tensor of size $U \times V \times 256$. Each grid cell $(u,v)$ stores a 256-d feature vector and corresponds to a 2cm$\times$2cm area on the floor-plan, which is the same spatial resolution as the segmentation $S$. 
The memory must be updated at each time step to incorporate new observations. 
Specifically, given the current memory $M^{(t-1)}$ and a new observation $F^{(t)}_{u,v}$ for cell $u,v$, 
we compute $M^{(t)}_{u,v} = GRU(F^{(t)}_{u,v}, M^{(t-1)}_{u,v})$ where $M^{(t-1)}_{u,v}$ is the hidden state and $F^{(t)}_{u,v}$ the input for the GRU. 
This GRU can learn to accumulate incoming observations. Notice that the GRU parameters are shared spatially, \ie for all $(u,v)$. 
Importantly, this independent updating of modified memory cells 
(as opposed to something like a ConvGRU) ensures that observations only affect local regions of the memory -- keeping previously observed areas stable.

\xhdr{\decoder.} The memory tensor $M^{(t)}$ is used to decode a top-down semantic segmentation map. 
We use a simple architecture consisting of five convolutional layers with batch norm and ReLU activations. As the memory $M$ and segmentation $S$ 
are the same spatial resolution, no learned upsampling or downsampling is involved in this decoding. 

Together, these modules form \ouralg and implement the basic principle of `encode pixels, project features to a spatial memory, decode labels'.
Notice that all modules and thus the entire architecture is end-to-end differentiable. 


\section{Matterport Semantic-Map Dataset}\label{sec:dataset}

For our experimental evaluation, we need 3D environments for an agent to traverse that have dense semantic segmentations. 
While our extension experiments involve agent-driven navigation, 
our core task of semantic mapping is defined \wrt a fixed trajectory provided to the agent as input. 
The more challenging task of simultaneous semantic mapping and goal-driven navigation is left for future work. 

Given this fixed-trajectory setting, our task has the input (but not output) specification of video segmentation -- both taking a sequence of input images along a trajectory. 
We choose the Matterport3D scans~\cite{chang2017matterport3d} with the Habitat simulator~\cite{habitat19iccv} 
over video segmentation datasets for a number of reasons -- 
Matterport3D provides semantic annotations in 3D (as opposed to 2D annotations in video datasets), 
the spaces are large enough to allow multi-room traversal by the agent 
(as opposed to \cite{dai2017scannet, Silberman:ECCV12}),
and the use of a 3D simulator (as opposed to \cite{Geiger2012CVPR,cordts2016cityscapes}) allows us render RGB-D from any viewpoint, create top-down semantic annotations, 
and study embodied AI applications in the same environments. 

\begin{figure*}[t]
\centering


\includegraphics[width=0.95\textwidth]{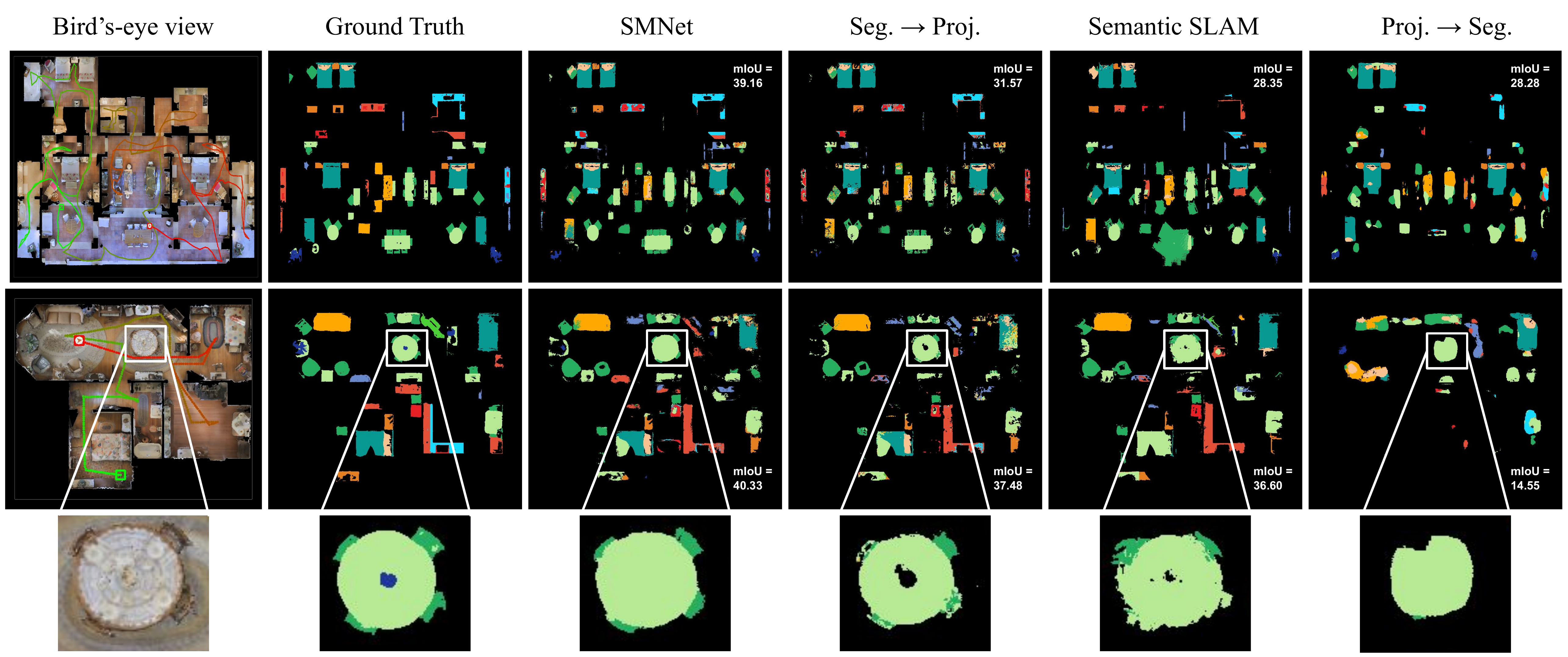} 

\begin{tabular}{r r r r r r r}
void \textcolor{void}{\rule{0.75cm}{0.25cm}} \hspace{0.5em} &
shelving \textcolor{shelving}{\rule{0.75cm}{0.25cm}} \hspace{0.5em} &
dresser \textcolor{chestofdrawers}{\rule{0.75cm}{0.25cm}} \hspace{0.5em} &
bed \textcolor{bed}{\rule{0.75cm}{0.25cm}} \hspace{0.5em} &
cushion \textcolor{cushion}{\rule{0.75cm}{0.25cm}} \hspace{0.5em} &
fireplace \textcolor{fireplace}{\rule{0.75cm}{0.25cm}} \hspace{0.5em} &
sofa \textcolor{sofa}{\rule{0.75cm}{0.25cm}} \hspace{0.5em} \\ & 
table \textcolor{table}{\rule{0.75cm}{0.25cm}} \hspace{0.5em} &
chair \textcolor{chair}{\rule{0.75cm}{0.25cm}} \hspace{0.5em} &
cabinet \textcolor{cabinet}{\rule{0.75cm}{0.25cm}} \hspace{0.5em} &
plant \textcolor{plant}{\rule{0.75cm}{0.25cm}} \hspace{0.5em} &
counter \textcolor{counter}{\rule{0.75cm}{0.25cm}} \hspace{0.5em} & sink \textcolor{sink}{\rule{0.75cm}{0.25cm}} \hspace{0.5em}~\\
\end{tabular}
\caption{Example semantic segmentation predictions. \ouralg makes cleaner and more accurate predictions than the baseline approaches.}
\label{fig:results_visualization}
\end{figure*}

\xhdr{Matterport3D Environments.} 
Matterport3D dataset \cite{chang2017matterport3d} contains reconstructed 3D meshes of 90 indoor environments (homes, offices, churches). 
These meshes are densely annotated with 40 object categories. Many of these are rare or would appear as thin lines in a top-down view (\eg walls and curtains); 
we focus on the 12 most common object categories: 
\texttt{chair, table, cushion, cabinet, shelving, sink, dresser, plant, bed, sofa, counter, fireplace} (sorted in descending order by number of object instances). 
We treat all other classes and background pixels as \texttt{void} class. 
We divide multi-story environments in Matterport3D into separate floors by manually refining floor dividers present in the meta-data. 
This is not always possible given a single dividing plane (\eg split level homes), resulting in inaccurate top-down maps -- we discard such environments. 
Utilizing the same data split as \cite{wijmans2019embodied}, we keep 85 unique floors in our dataset: 61 for training, 7 for validation, and 17 for testing. See supplement for these splits.

\xhdr{Ground-Truth Top-down Semantic Segmentations.} 
To supervise our model, we need access to ground-truth top-down semantic maps from these environments. 
These are created by applying an orthographic projection for the 3D mesh annotations in a similar manner to \eqnref{eq:mapper} (right). In this process we only project vertices labeled with one of the 12 kept object categories. The resulting ground-truth top-down semantic maps are free from occlusions caused by non-target objects. There will be cases where from the egocentric view the agent won't be able to visualize the object entirely either because it is occluded or the object is too high (wall cabinet). In table \ref{tab:results} we report numbers on the Seg. GT $\rightarrow$ Proj. experiment where the agent projects egocentric ground-truth semantic labels. This will account for such occlusion and set an upper-bound to our experiments.

\xhdr{Modal Maps and Viewing Frustum.}
We perform \emph{modal} top-down semantic segmentation (as opposed to \emph{amodal}). Specifically, the agent receives 
supervision on map cells it has \emph{actually observed}; it is not evaluated on hallucinating unseen regions. 
We do this by projecting the viewing frustum (\ie region the agent can currently see) to the floor-plan at each navigation step. 
We can then keep track of which regions have been observed during a trajectory. 

\begin{table*}[t]
\centering
\renewcommand{\arraystretch}{1.15}
\resizebox{0.95\textwidth}{!}{
\begin{tabular}{ l c c c c s s c c c c s s}
\toprule
& & \multicolumn{5}{c}{Matteport3D (test)} &  & \multicolumn{5}{c}{Replica}\\
\cline{3-7}\cline{9-13}
& & Acc & mRecall & mPrecision & \multicolumn{1}{c}{mIoU} & \multicolumn{1}{c}{mBF1} & & Acc & mRecall & mPrecision & \multicolumn{1}{c}{mIoU} & \multicolumn{1}{c}{mBF1}\\ 
\midrule
Seg. GT $\rightarrow$ Proj. & & 89.49 {\scriptsize $\pm$ 0.09} & 73.73 {\scriptsize $\pm$ 0.06} & 74.58 {\scriptsize $\pm$ 0.10} & 59.73 {\scriptsize $\pm$ 0.09} & 54.05 {\scriptsize $\pm$ 0.11} & & 96.83 {\scriptsize $\pm$ 0.07} & 83.84 {\scriptsize $\pm$ 0.05} & 94.05 {\scriptsize $\pm$ 0.06} & 79.76 {\scriptsize $\pm$ 0.07} & 86.89 {\scriptsize $\pm$ 0.04} \\
\midrule
Proj. $\rightarrow$ Seg. & & 83.18 {\scriptsize $\pm$ 0.07} & 27.32 {\scriptsize $\pm$ 0.08} & 35.30 {\scriptsize $\pm$ 0.13} & 19.96 {\scriptsize $\pm$ 0.07} & 17.33 {\scriptsize $\pm$ 0.08} & &
81.25 {\scriptsize $\pm$ 0.09} & 26.64 {\scriptsize $\pm$ 0.12} & 41.50 {\scriptsize $\pm$ 0.12} & 20.06 {\scriptsize $\pm$ 0.09} & 19.08 {\scriptsize $\pm$ 0.12} \\

Seg. $\rightarrow$ Proj. & & 88.06 {\scriptsize $\pm$ 0.07} & 40.53 {\scriptsize $\pm$ 0.09} & \textbf{58.92 {\scriptsize $\pm$ 0.11}} & 32.76 {\scriptsize $\pm$ 0.07} & 33.21 {\scriptsize $\pm$ 0.08} & & 88.61 {\scriptsize $\pm$ 0.09} & 48.11 {\scriptsize $\pm$ 0.09} & \textbf{65.20 {\scriptsize $\pm$ 0.11}} & 40.77 {\scriptsize $\pm$ 0.09} & 45.86 {\scriptsize $\pm$ 0.12}\\
Semantic SLAM & & 85.17 {\scriptsize $\pm$ 0.08} & 37.51 {\scriptsize $\pm$ 0.09} & 51.54 {\scriptsize $\pm$ 0.15} & 28.11 {\scriptsize $\pm$ 0.08} & 31.05 {\scriptsize $\pm$ 0.12} & & 88.30 {\scriptsize $\pm$ 0.09} & 45.80 {\scriptsize $\pm$ 0.08} & 62.41 {\scriptsize $\pm$ 0.12} & 37.99 {\scriptsize $\pm$ 0.09} & \textbf{46.71 {\scriptsize $\pm$ 0.10}} \\
\midrule
\ouralg & & \textbf{88.14 {\scriptsize $\pm$ 0.09}} & \textbf{47.49 {\scriptsize $\pm$ 0.11}} & 58.27 {\scriptsize $\pm$ 0.11} & \textbf{36.77 {\scriptsize $\pm$ 0.09}} & \textbf{37.02 {\scriptsize $\pm$ 0.09}}& & \textbf{89.26 {\scriptsize $\pm$ 0.10}} & \textbf{53.37 {\scriptsize $\pm$ 0.12}} & 64.81 {\scriptsize $\pm$ 0.09} & \textbf{43.12 {\scriptsize $\pm$ 0.10}} & 45.18 {\scriptsize $\pm$ 0.14}\\
\bottomrule
\end{tabular}}\\[2pt]
\caption{Results on top-down semantic segmentation on the Matterport3D and Replica datasets. Models have not been trained on Replica and those results are purely transfer experiments. \ouralg outperforms the baselines on mIoU and mBF1 for Matterport3D and mIoU in Replica. }
\label{tab:results}
\end{table*}

\xhdr{Navigation Paths.} We assume that an agent's path through the environment is provided by some external policy -- \eg a goal-oriented path or a general exploration 
policy -- and that we are constructing the map and memory opportunistically from this experience. 
To simulate this behavior, we manually record a navigation path through each floor using the Habitat simulator~\cite{habitat19iccv}. 
The action space is move forward 10cm,
and rotate left or right 9$^\circ$ . To encourage trajectories with high environment coverage, 
our human navigation interface included the top-down RGB map with agent position drawn. 
On average, agents move 2500 steps in each environment. Note that this is an order of magnitude longer than most navigation trajectories in contemporary 
works~\cite{habitat19iccv, wijmans2019decentralized,habitatsim2real19arxiv, gordon2018iqa, chaplot2020learning}

\xhdr{Training Samples.} To train our model, we consider 20-step navigation segments from these trajectories. Starting from a random location on the trajectory, we step the agent forward 20 steps along it, capturing the corresponding viewpoints to mask the top-down semantic map. We generate 50 examples for each environment leading to 3050/350 train/val training samples. This both greatly increases the number of training instances and increases training speed by requiring a smaller semantic memory tensor. At evaluation/testing time, the agent 
builds the map from the entire trajectory.


\section{Semantic Mapping Experiments}

\xhdr{Baselines.} As depicted in \figref{fig:related}, there exists a spectrum of methodologies for our task based on what is being projected from egocentric observations to the top-down map -- pixels, features, or labels. Our approach stakes a middle-ground on this spectrum -- projecting egocentric features. We compare with approaches at either end and existing work:

\begin{compactitem}[\hspace{0pt}--]


\item \textbf{Project $\rightarrow$ Segment.} As agents traverse the scene, the observed RGB pixels are projected to the top-down map using our mapper architecture -- resulting in a top-down RGB image of the environment. We train a model similar to RedNet~\cite{jiang2018rednet} to decode the semantic map directly from this top-down RGB image. 

\item\textbf{Segment $\rightarrow$ Project.} At the other extreme, agents perform semantic segmentation on each egocentric frame and then project the resulting labels using our mapper architecture to create the top-down segmentation. The produced top-down semantic maps are post-processed using a median filter (3$\times$3) to reduce the label splatter noise caused by egocentric prediction errors around object boundaries. In addition, we found experimentally that down-sampling the egocentric resolution from (480$\times$640) to (120$\times$160) helps reducing the impact of egocentric errors on the top-down maps and leads to best performances for this baseline. Any missed pixels in the observed area of the top-down semantic map caused by this down-sampling is filled using median filtering. We fine-tune a RedNet~\cite{jiang2018rednet} model for this task. We also present an oracle baseline that projects ground-truth segmentations -- \textbf{Segment GT $\rightarrow$ Project.} This experiment sets an upper-bound of performances (perfect predictions being not possible due to occlusions)

\item \textbf{Semantic SLAM.} We use VoxBlox++, an off-the-shelf implementation of semantic SLAM~\cite{grinvald2019volumetric} where we replaced the object detection module with our pre-trained RedNet model plus a hand-crafted instance segmentation applied on-top of the semantic predictions (connected components). The algorithm takes as input RGB-D frames and simultaneously estimates agent's pose 
and constructs a point cloud of semantically labelled points. We project the point cloud to a top-down segmentation using the same mapping functions we described in Sec.~\ref{sec:memory}. For fairness, we provide the ground truth pose to \cite{grinvald2019volumetric} at each time step. 

\end{compactitem}
In all experiments, detections 50cm over the agent's camera position are discarded. This prevents detecting ceilings. 

\xhdr{Implementation Details} 
We pretrain two RedNet \cite{jiang2018rednet} models for semantic segmentation in our setting -- one from egocentric RGB-D (Segment$\rightarrow$Project) and another from top-down RGB alone (Project$\rightarrow$Segment). \ouralg is initialized with the encoder from the egocentric RedNet. We use a single-layer GRU to update the spatial memory. \ouralg is  trained end-to-end under cross-entropy loss using SGD with learning rate $1\mathrm{e}{-4}$, momentum $0.9$, weight decay $4\mathrm{e}{-4}$, and batch size 8 across 8 Titan XPs. Training took 2-3 days. Back propagation is applied after 20 steps.

\xhdr{Evaluation Metrics.} We report the entire range of evaluation metrics for semantic segmentation: 
(a) the overall pixel-wise labeling accuracy (\textbf{Acc}), 
(b) the average of pixel recall or precision scores for each class (\textbf{mRecall/mPrecision}), 
(c) the average of the intersection-over-union score of all object categories (\textbf{mIoU}), and 
(d) the average of the boundary F1 score of all object categories (\textbf{mBF1}). mBF1 is contour-based metric defined in \cite{csurka2013good}. mIoU and mBF1 serve as our primary metrics. 

\begin{figure*}[t]
\centering
\setlength{\tabcolsep}{2pt}
\hspace{-1em}
\resizebox{0.96\textwidth}{!}{
\begin{tabular}{c c c c}
\multicolumn{2}{c}{Goal: Go to the closest bed} & \multicolumn{2}{c}{Goal: Go to the closest chair} \\
\includegraphics[width=0.24\textwidth,height=1.2in]{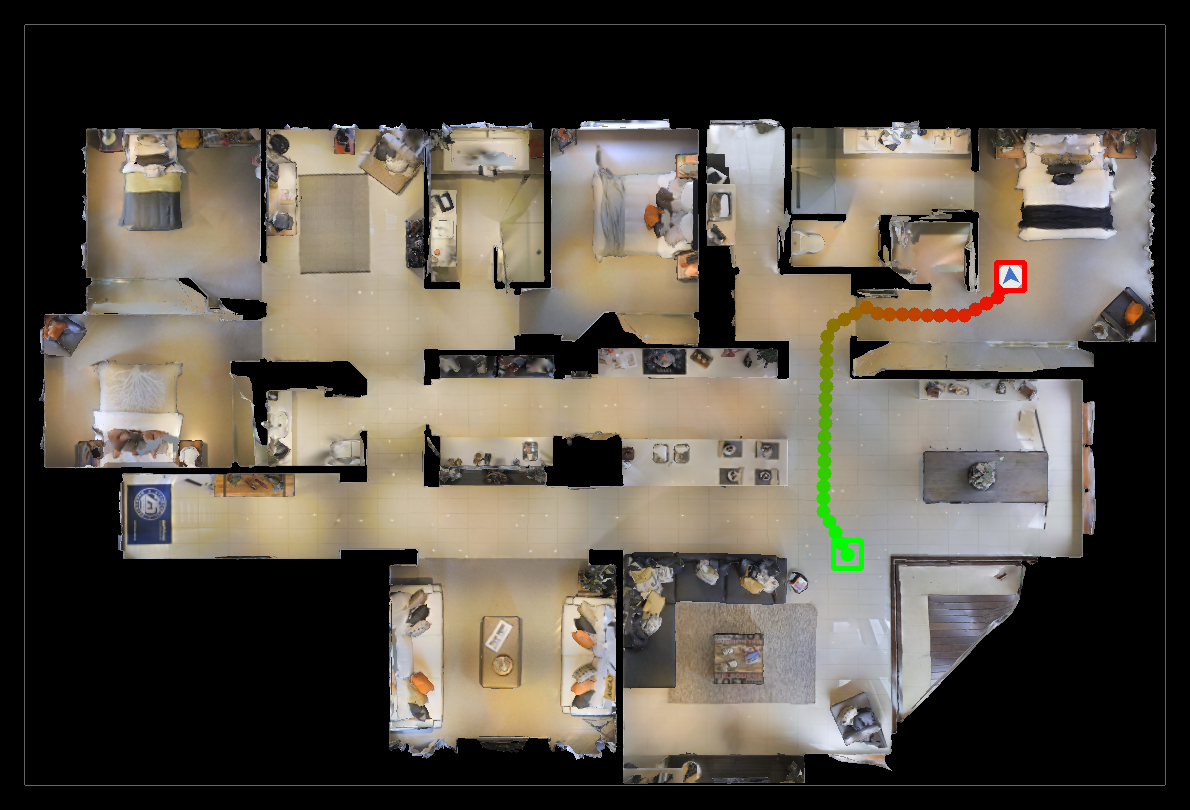} &
\includegraphics[width=0.24\textwidth,height=1.2in]{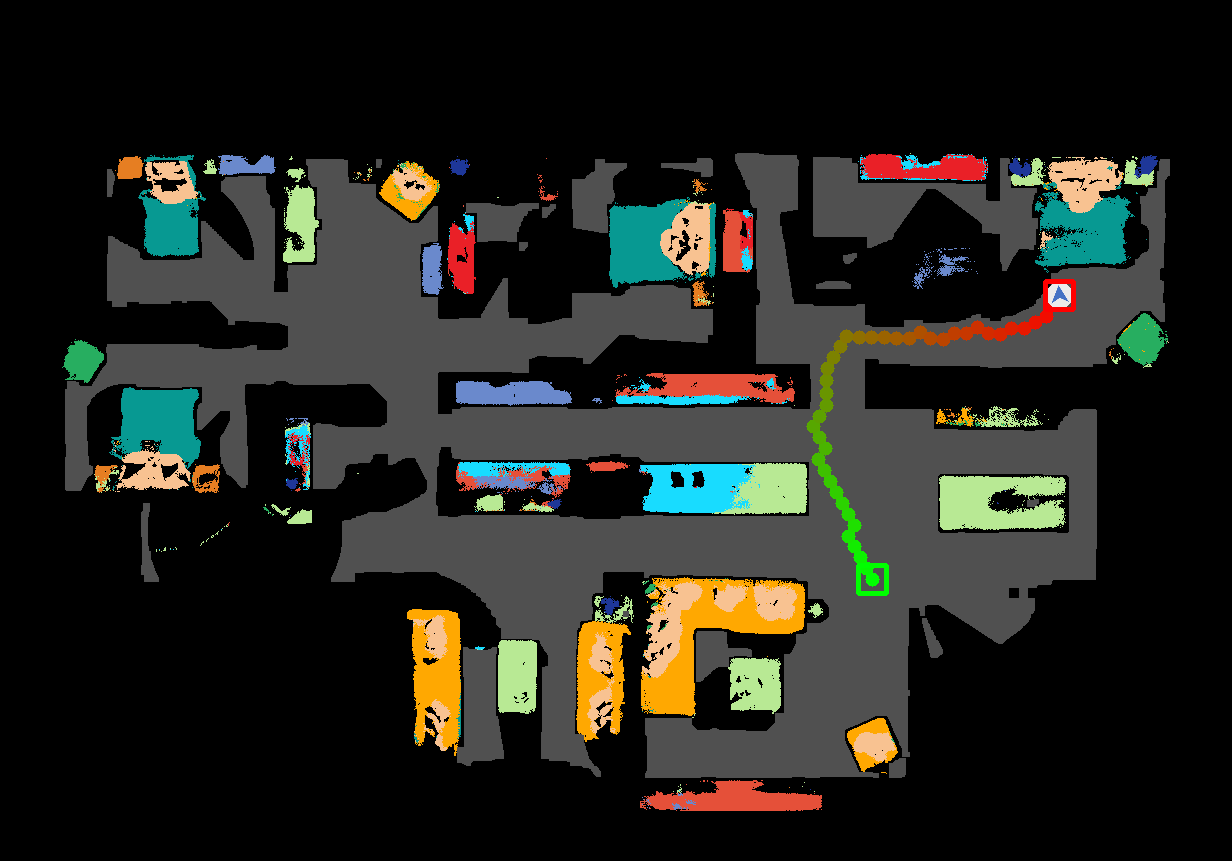} \hspace{0.5em} &
\includegraphics[width=0.24\textwidth,height=1.2in]{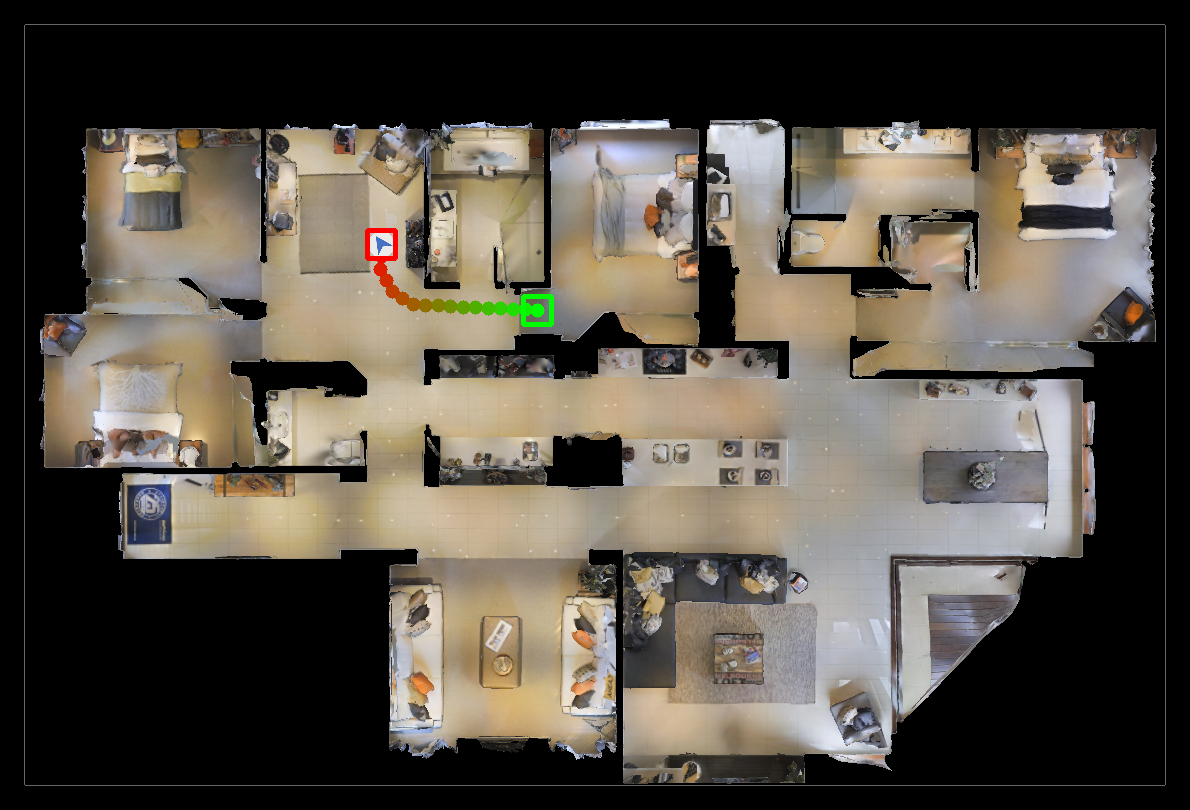} &
\includegraphics[width=0.24\textwidth,height=1.2in]{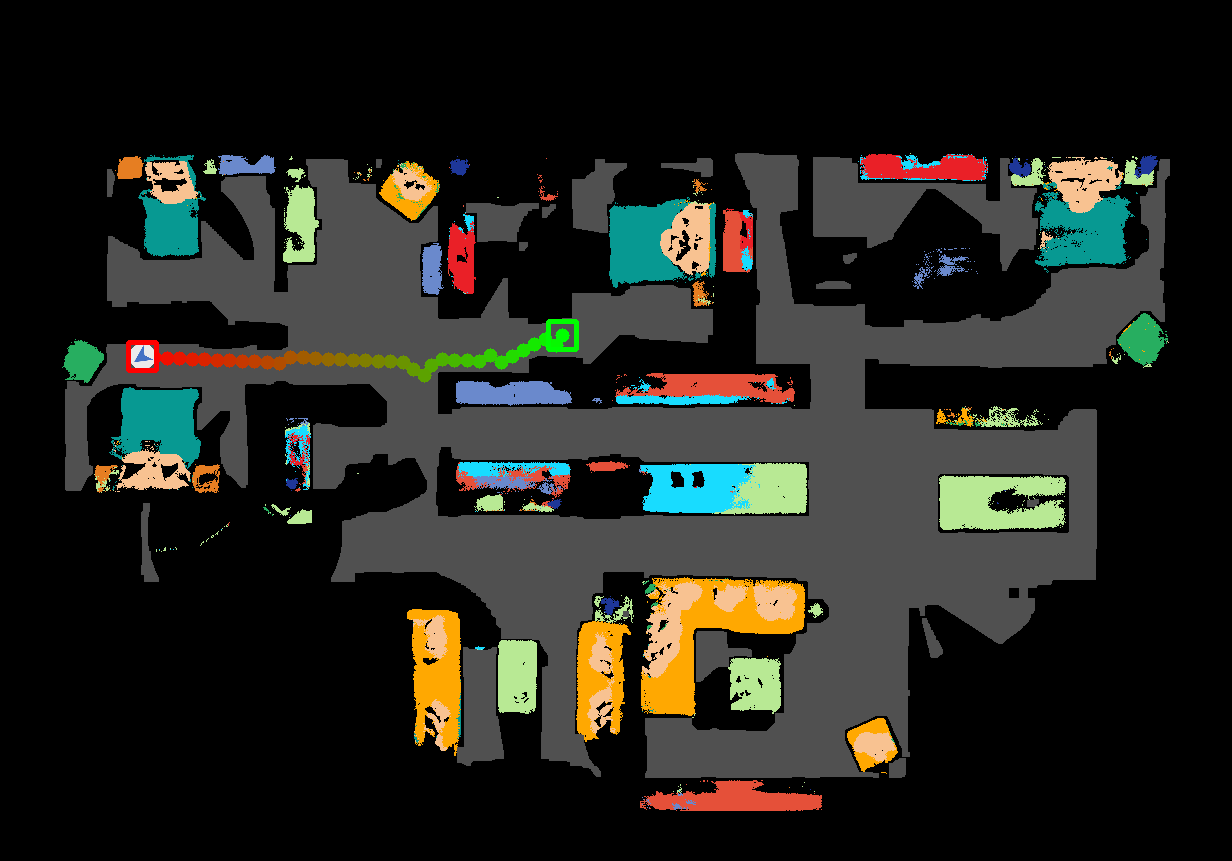} \\
\footnotesize Ground Truth & \footnotesize \ouralg \hspace{0.4em} & \footnotesize Ground Truth & \footnotesize \ouralg \\
\end{tabular}}
\caption{Object Navigation: Visualization of paths found by A* using \ouralg maps. Green and red squares indicate agent's starting and stopping locations. The grey color represents the floor pixels. 
The left example shows a case of success with high SPL $=0.8276$ and the example on the right shows a case of success with low SPL$=0.4282$. }
\label{fig:results_visualization_objnav}
\end{figure*}

\begin{figure}[t]
\centering
\vspace{1em}
\setlength{\tabcolsep}{2pt}
\hspace{-1em}
\begin{tabular}{c c c }
\footnotesize Bird's-eye View & \footnotesize Ground Truth & \footnotesize \ouralg  \\
\includegraphics[width=0.33\linewidth,height=1.1in]{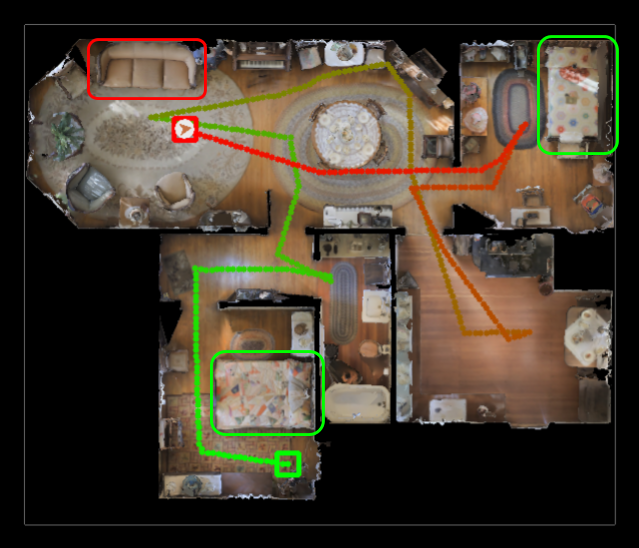} &
\includegraphics[width=0.33\linewidth,height=1.1in]{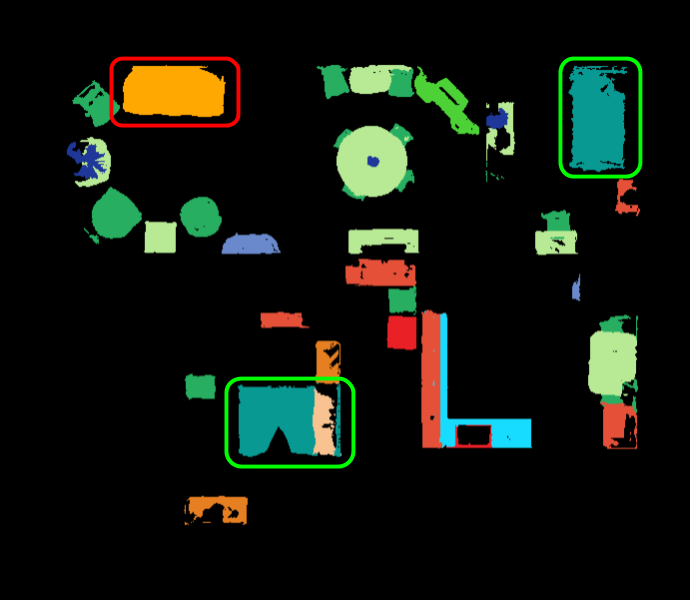} &
\includegraphics[width=0.33\linewidth,height=1.1in]{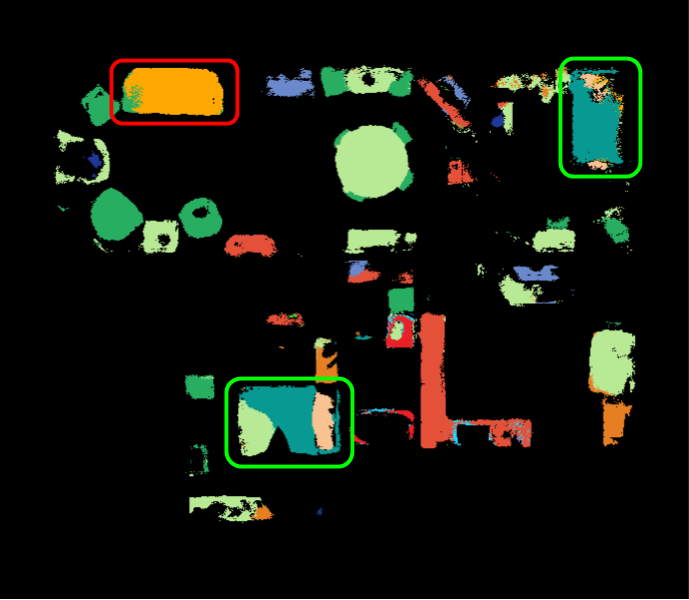}\\
\multicolumn{3}{c}{How many \setulcolor{red} \ul{sofas} are there?  GT: 1  --  Pred: 1} \\
\multicolumn{3}{c}{How many \setulcolor{green} \ul{beds} are there?  GT: 2  --  Pred: 3} \\
\end{tabular}
\vspace{1em}
\caption{Visualizations of MemoryQA}
\label{fig:results_visualization_memqa}
\end{figure}

\xhdr{Results.} 
\tableref{tab:results}(left) shows a summary of the results with bootstrapped standard error (see supplement for category-level breakdowns). \figref{fig:results_visualization} shows qualitative results. Project{$\rightarrow$}Segment achieves low performance (mBF1 17.33, mIoU 19.96) compared to the approaches that operate on egocentric images prior to projection (either via segmentation or feature extraction). This suggests details lost in the top-down view are important for disambiguating objects -- \eg the chairs at the table in \figref{fig:results_visualization} (bottom) are difficult to see in the top-down RGB and are completely lost by this approach. Segment{$\rightarrow$}Project performs significantly better (mBF1 33.21, mIoU 32.76), but faces a problem with errors in the egocentric predictions resulting in noise in the top-down map.  
Semantic SLAM (VoxBlobx++) performs worse than the Segment{$\rightarrow$}Project baseline (mBF1 31.05, mIoU 28.11).
VoxBlox++ follows a segment-then-project paradigm, making it prone to the same errors as the Segment $\rightarrow$ Project baseline. In addition, the data association module of VoxBlox++ will sometimes group objects of different categories (\eg the two bottom chairs are grouped with the table in \figref{fig:results_visualization}).
As our approach reasons over a spatial memory tensor, it can reason about multiple observations of the same point -- achieving mBF1 37.02, and mIoU 36.77. 
The Segment GT {$\rightarrow$}Project experiment sets an upper-bound of mBF1 54.05, and mIoU 59.73.
We also evaluated SMNet on the replica dataset \cite{straub2019replica}. \tableref{tab:results} (right) shows a summary of the results with bootstrapped standard error.
Similarly, \ouralg performs best on the mIoU metric at 43.12.
These results demonstrate that an approach which interleaves projective geometry and learning can provide more robust allocentric semantic representations.


\section{Re-using Maps for Downstream Tasks}
\label{sec:embodied}

The map and spatio-semantic allocentric representation our method constructs while exploring an environment provide a rich description of the space. In this section, we explore proof-of-concepts for various downstream embodied AI tasks based on these representations.

\xhdr{Object Navigation.} A natural extension is navigating to specific objects, or ObjectNav for short. In ObjectNav, an agent is randomly initialized in a scene and tasked to navigate to an instance of a given object class as quickly as possible \cite{habitat19iccv}. In the standard setting, the environment is novel; however, we consider a pre-exploration setting where the agent first traverses the environment to construct a top-down semantic map. In parallel we compute a top-down map of heights and use it to compute a free space map of the environment. We opt for an open loop planing strategy by running A* search (with a Euclidean heuristic) on the free space map combined with the semantic map to find a path from the start location to the nearest target object instance and then run the trajectory in the Habitat simulator \cite{habitat19iccv}. 
We evaluated this strategy on the validation set of the ObjectNav Habitat challenge \cite{habitat_challenge20}, agents are able to achieve a success rate of $9.658 \%$, with SPL of $5.714 \%$, soft SPL of $ 8.702\%$ and average distance to the target of $7.31576m$. Note that $26 \%$ of the episodes in this set are targeting object categories falling outside of our list of object classes, we consider those as failure. 
The evaluation metrics limited to episodes targeting objects in our list of classes are: success rate of $13.070 \%$, with SPL of $7.733 \%$, soft SPL of $11.777\%$ and average distance to the target of $6.70981m$.
These results are in the same order of magnitude as the state-of-the-art methods submitted to the Habitat Challenge~\cite{habitat_challenge20},
suggesting that the memory tensor contains useful spatial and semantic information in this pre-exploration setting. 
Experimentally we found that inaccuracies in the free space map computation and objects misclassification in the top-down semantic map are the two major sources of error. While the latter is harder to cope with, the former can be limited by extended \ouralg to predict free space.
\figref{fig:results_visualization_objnav} shows qualitative results of two successful examples -- start locations are shown as green squares with trajectory transitioning to red until terminating. 
Using the predicted semantic maps provides interpretability -- 
when the navigation fails, we can know why. On the example on the right in  \figref{fig:results_visualization_objnav} the chair at the top has been mislabeled as sofa, thus leading the agent to the second closest chair slightly on the left.


\xhdr{Question Answering.} 
We also consider an embodied question answering~\cite{das2018embodied} task where agents are asked questions about the environment. 
Again considering a pre-exploration setting, the agent first navigates the environment on a fixed trajectory to generate the spatial memory tensor. 
We consider counting questions (\eg\myquote{How many beds are there?}) 
and design a decoder directly from the spatial memory. The decoder outputs the number of instances detected per object category for a given memory input. 
We train this decoder using 5m x 5m memory samples. We design this task as a classification problem with 21 classes corresponding to values ranging from 0 to 19 and 20+.
When testing on larger environments, we apply this decoder using a sliding window over the full memory -- accumulating counts. We compare our approach to a `prior' baseline that answers 
with the most frequent answer in the training set. Our approach outperforms this baseline across the board: $27.78$\% \vs $20.83$\% on accuracy, 
$13.19$\% \vs $9.18\%$ class-balanced accuracy and $5.35$ \vs $6.98$ on RMSE. 

\section{Conclusion}
\label{sec:conclusion}
Taken holistically, our results show \ouralg is able to outperform competitive baselines in constructing semantic maps, and spatio-semantic representations built 
show promise on downstream tasks. Note that the specific sub-task of counting instances highlights a limitation in our current problem setup -- 
using semantic segmentation does not preserve instance information. The generalization to producing top-down instance segmentation maps is an interesting avenue for future work. 


\bibliography{bib}

\newpage

\input{supplementary}

\end{document}

%% file: supplementary.tex
\section{Appendix}


\subsection{Data splits of the Dataset}
We use the same data split as Wijmans~\etal~\cite{wijmans2019embodied}, and provide a list of environment names in the Matterport3D dataset~\cite{chang2017matterport3d} in Table~\ref{tab:envs}. There are 85 unique floors in our dataset, 61 for training, 7 for validations, and 17 for testing.

We also evaluated \ouralg on the Replica dataset~\cite{straub2019replica}. Table~\ref{tab:envs_replica} lists the 17 environments used for evaluation.

\begin{table*}[!h]
\centering
\begin{tabular}{llll}
\rowcolor{gray!10} \multicolumn{4}{c}{Train Environments} \\
ZMojNkEp431\_0 & ZMojNkEp431\_1 & aayBHfsNo7d\_0 & aayBHfsNo7d\_1 \\
ac26ZMwG7aT\_0 & cV4RVeZvu5T\_0 & cV4RVeZvu5T\_1 & e9zR4mvMWw7\_1 \\
e9zR4mvMWw7\_2 & i5noydFURQK\_0 & i5noydFURQK\_1 & jh4fc5c5qoQ\_0 \\
mJXqzFtmKg4\_0 & p5wJjkQkbXX\_1 & r1Q1Z4BcV1o\_0 & r47D5H71a5s\_0 \\
rPc6DW4iMge\_0 & rPc6DW4iMge\_1 & sT4fr6TAbpF\_0 & uNb9QFRL6hY\_0 \\
17DRP5sb8fy\_0 & 1LXtFkjw3qL\_0 & 1LXtFkjw3qL\_1 & 1LXtFkjw3qL\_2 \\
1pXnuDYAj8r\_0 & 1pXnuDYAj8r\_1 & 29hnd4uzFmX\_0 & 29hnd4uzFmX\_1 \\
2n8kARJN3HM\_0 & 2n8kARJN3HM\_1 & 5LpN3gDmAk7\_0 & 5q7pvUzZiYa\_0 \\
5q7pvUzZiYa\_1 & 5q7pvUzZiYa\_2 & 759xd9YjKW5\_0 & 759xd9YjKW5\_1 \\
82sE5b5pLXE\_0 & 8WUmhLawc2A\_0 & B6ByNegPMKs\_0 & E9uDoFAP3SH\_0 \\
E9uDoFAP3SH\_1 & EDJbREhghzL\_0 & GdvgFV5R1Z5\_0 & HxpKQynjfin\_0 \\
JF19kD82Mey\_0 & JF19kD82Mey\_1 & JeFG25nYj2p\_0 & JmbYfDe2QKZ\_0 \\
JmbYfDe2QKZ\_1 & Pm6F8kyY3z2\_1 & PuKPg4mmafe\_0 & S9hNv5qa7GM\_0 \\
S9hNv5qa7GM\_1 & ULsKaCPVFJR\_0 & ULsKaCPVFJR\_1 & Uxmj2M2itWa\_0 \\
VLzqgDo317F\_1 & VVfe2KiqLaN\_0 & VVfe2KiqLaN\_1 & Vvot9Ly1tCj\_0 \\
ur6pFq6Qu1A\_0 & & & \\
\rowcolor{gray!10} \multicolumn{4}{c}{Val Environments} \\
Z6MFQCViBuw\_0 & x8F5xyUWy9e\_0 & zsNo4HB9uLZ\_0 & 8194nk5LbLH\_0 \\
QUCTc6BB5sX\_0 & QUCTc6BB5sX\_1 & X7HyMhZNoso\_0 \\
\rowcolor{gray!10} \multicolumn{4}{c}{Test Environments} \\
YFuZgdQ5vWj\_0 & YFuZgdQ5vWj\_1 & YVUC4YcDtcY\_0 & jtcxE69GiFV\_0 \\
q9vSo1VnCiC\_0 & rqfALeAoiTq\_1 & rqfALeAoiTq\_2 & wc2JMjhGNzB\_1 \\
2t7WUuJeko7\_0 & 5ZKStnWn8Zo\_0 & 5ZKStnWn8Zo\_1 & ARNzJeq3xxb\_0 \\
RPmz2sHmrrY\_0 & UwV83HsGsw3\_1 & Vt2qJdWjCF2\_0 & WYY7iVyf5p8\_1 \\
UwV83HsGsw3\_0

\end{tabular}
\caption{Train/val/test environments for Matterport3D scenes~\cite{chang2017matterport3d} in our dataset}
\label{tab:envs}
\end{table*}

\begin{table*}[!h]
\centering
\begin{tabular}{llll}
\rowcolor{gray!10} \multicolumn{4}{c}{Test Environments} \\
apartment\_0 & apartment\_1 & apartment\_2 & frl\_apartment\_0 \\
frl\_apartment\_1 & frl\_apartment\_2 & frl\_apartment\_3 & frl\_apartment\_4 \\
frl\_apartment\_5 & hotel\_0 & office\_0 & office\_1 \\
office\_2 & office\_3 & office\_4 & room\_0 \\
room\_1
\end{tabular}
\caption{Test environments for Replica scenes~\cite{straub2019replica}}
\label{tab:envs_replica}
\end{table*}


\subsection{Evaluations for each object category}
In Table~\ref{tab:breakdown_results}, we perform evaluations for each object category in the test set of the Matterport3D dataset~\cite{chang2017matterport3d}.  \ouralg consistently out-perform baseline approaches for most object categories.
We also compute evaluation metrics for each object category on the Replica dataset~\cite{straub2019replica} in Table~\ref{tab:breakdown_results_replica}. Here also  \ouralg consistently out-perform baseline approaches for most object categories.

\begin{table*}[!h]
 
\centering
\resizebox{\textwidth}{!}{
\setlength{\tabcolsep}{3pt}
\centering
\begin{tabular}{c c c c c c c c c c c c c c c c c c c c c}
\toprule
\multicolumn{2}{c}{} & \multicolumn{5}{c}{Recall} & \multicolumn{5}{c}{Precision} & \multicolumn{5}{c}{IoU} & \multicolumn{4}{c}{BF1} \\
\cmidrule(r){3-6} 
\cmidrule(r){8-11} 
\cmidrule(r){13-16} 
\cmidrule(r){18-21} 
  & & \scriptsize \shortstack{Segment $\rightarrow$ \\ Project} & \scriptsize \shortstack{Project $\rightarrow$ \\ Segment} & \scriptsize semantic SLAM &  \scriptsize \ouralg 
  & & \scriptsize \shortstack{Segment $\rightarrow$ \\ Project} & \scriptsize \shortstack{Project $\rightarrow$ \\ Segment} & \scriptsize semantic SLAM &  \scriptsize \ouralg 
  & & \scriptsize \shortstack{Segment $\rightarrow$ \\ Project} & \scriptsize \shortstack{Project $\rightarrow$ \\ Segment} & \scriptsize semantic SLAM &  \scriptsize \ouralg 
  & & \scriptsize \shortstack{Segment $\rightarrow$ \\ Project} & \scriptsize \shortstack{Project $\rightarrow$ \\ Segment} & \scriptsize semantic SLAM &  \scriptsize \ouralg  \\

\midrule
void  & & 98.22 {\scriptsize $\pm$ 0.02} & 95.15 {\scriptsize $\pm$ 0.02} & 95.11 {\scriptsize $\pm$ 0.08} & 96.27 {\scriptsize $\pm$ 0.03} &  & 91.35 {\scriptsize $\pm$ 0.04} & 89.57 {\scriptsize $\pm$ 0.06} & 91.46 {\scriptsize $\pm$ 0.04} & 93.42 {\scriptsize $\pm$ 0.02} &  & 89.86 {\scriptsize $\pm$ 0.04} & 85.66 {\scriptsize $\pm$ 0.05} & 87.38 {\scriptsize $\pm$ 0.11} & 90.16 {\scriptsize $\pm$ 0.04} &  & - & - & - \\
shelving  & & 17.86 {\scriptsize $\pm$ 0.20} & 5.07 {\scriptsize $\pm$ 0.13} & 17.96 {\scriptsize $\pm$ 0.17} & 21.33 {\scriptsize $\pm$ 0.24} &  & 34.29 {\scriptsize $\pm$ 0.32} & 13.24 {\scriptsize $\pm$ 0.30} & 44.38 {\scriptsize $\pm$ 0.43} & 36.24 {\scriptsize $\pm$ 0.22} &  & 13.15 {\scriptsize $\pm$ 0.13} & 3.78 {\scriptsize $\pm$ 0.09} & 14.50 {\scriptsize $\pm$ 0.13} & 15.34 {\scriptsize $\pm$ 0.14} &  & 21.76 {\scriptsize $\pm$ 0.15} & 7.34 {\scriptsize $\pm$ 0.11} & 22.05 {\scriptsize $\pm$ 0.24} & 24.56 {\scriptsize $\pm$ 0.18} \\
dresser  & & 19.57 {\scriptsize $\pm$ 0.12} & 16.57 {\scriptsize $\pm$ 0.24} & 21.90 {\scriptsize $\pm$ 0.27} & 28.03 {\scriptsize $\pm$ 0.23} &  & 42.95 {\scriptsize $\pm$ 0.47} & 32.14 {\scriptsize $\pm$ 0.45} & 43.08 {\scriptsize $\pm$ 0.51} & 47.82 {\scriptsize $\pm$ 0.50} &  & 15.32 {\scriptsize $\pm$ 0.11} & 12.18 {\scriptsize $\pm$ 0.18} & 16.80 {\scriptsize $\pm$ 0.21} & 21.15 {\scriptsize $\pm$ 0.18} &  & 28.94 {\scriptsize $\pm$ 0.18} & 14.29 {\scriptsize $\pm$ 0.22} & 28.34 {\scriptsize $\pm$ 0.29} & 31.52 {\scriptsize $\pm$ 0.18} \\
bed  & & 69.52 {\scriptsize $\pm$ 0.27} & 68.84 {\scriptsize $\pm$ 0.35} & 57.45 {\scriptsize $\pm$ 0.36} & 71.73 {\scriptsize $\pm$ 0.20} &  & 83.69 {\scriptsize $\pm$ 0.18} & 52.01 {\scriptsize $\pm$ 0.46} & 72.33 {\scriptsize $\pm$ 0.30} & 85.53 {\scriptsize $\pm$ 0.20} &  & 61.13 {\scriptsize $\pm$ 0.23} & 41.94 {\scriptsize $\pm$ 0.36} & 47.03 {\scriptsize $\pm$ 0.30} & 63.91 {\scriptsize $\pm$ 0.19} &  & 51.19 {\scriptsize $\pm$ 0.22} & 33.49 {\scriptsize $\pm$ 0.26} & 48.99 {\scriptsize $\pm$ 0.25} & 54.57 {\scriptsize $\pm$ 0.23} \\
cushion  & & 70.71 {\scriptsize $\pm$ 0.21} & 34.81 {\scriptsize $\pm$ 0.27} & 46.89 {\scriptsize $\pm$ 0.20} & 78.25 {\scriptsize $\pm$ 0.24} &  & 58.95 {\scriptsize $\pm$ 0.31} & 43.76 {\scriptsize $\pm$ 0.34} & 42.06 {\scriptsize $\pm$ 0.41} & 56.83 {\scriptsize $\pm$ 0.27} &  & 47.21 {\scriptsize $\pm$ 0.22} & 24.02 {\scriptsize $\pm$ 0.20} & 28.20 {\scriptsize $\pm$ 0.21} & 48.95 {\scriptsize $\pm$ 0.22} &  & 50.91 {\scriptsize $\pm$ 0.23} & 33.55 {\scriptsize $\pm$ 0.19} & 43.14 {\scriptsize $\pm$ 0.22} & 53.01 {\scriptsize $\pm$ 0.22} \\
fireplace  & & 20.84 {\scriptsize $\pm$ 0.29} & 0.00 {\scriptsize $\pm$ 0.00} & 29.61 {\scriptsize $\pm$ 0.40} & 30.32 {\scriptsize $\pm$ 0.57} &  & 63.58 {\scriptsize $\pm$ 0.31} & 0.00 {\scriptsize $\pm$ 0.00} & 28.58 {\scriptsize $\pm$ 1.59} & 68.92 {\scriptsize $\pm$ 0.33} &  & 18.59 {\scriptsize $\pm$ 0.25} & 0.00 {\scriptsize $\pm$ 0.00} & 11.09 {\scriptsize $\pm$ 0.59} & 26.59 {\scriptsize $\pm$ 0.47} &  & 26.32 {\scriptsize $\pm$ 0.33} & 0.00 {\scriptsize $\pm$ 0.00} & 24.72 {\scriptsize $\pm$ 0.61} & 37.07 {\scriptsize $\pm$ 0.43} \\
sofa  & & 50.21 {\scriptsize $\pm$ 0.31} & 27.96 {\scriptsize $\pm$ 0.28} & 51.35 {\scriptsize $\pm$ 0.28} & 43.30 {\scriptsize $\pm$ 0.30} &  & 51.69 {\scriptsize $\pm$ 0.26} & 24.83 {\scriptsize $\pm$ 0.23} & 44.03 {\scriptsize $\pm$ 0.24} & 56.60 {\scriptsize $\pm$ 0.33} &  & 33.99 {\scriptsize $\pm$ 0.19} & 15.00 {\scriptsize $\pm$ 0.14} & 31.11 {\scriptsize $\pm$ 0.20} & 32.35 {\scriptsize $\pm$ 0.22} &  & 30.00 {\scriptsize $\pm$ 0.15} & 15.78 {\scriptsize $\pm$ 0.12} & 36.59 {\scriptsize $\pm$ 0.21} & 34.28 {\scriptsize $\pm$ 0.19} \\
table  & & 56.12 {\scriptsize $\pm$ 0.21} & 38.73 {\scriptsize $\pm$ 0.14} & 56.46 {\scriptsize $\pm$ 0.21} & 65.10 {\scriptsize $\pm$ 0.21} &  & 64.96 {\scriptsize $\pm$ 0.26} & 38.30 {\scriptsize $\pm$ 0.20} & 67.32 {\scriptsize $\pm$ 0.22} & 59.23 {\scriptsize $\pm$ 0.26} &  & 43.01 {\scriptsize $\pm$ 0.18} & 23.83 {\scriptsize $\pm$ 0.12} & 44.25 {\scriptsize $\pm$ 0.18} & 44.88 {\scriptsize $\pm$ 0.19} &  & 29.42 {\scriptsize $\pm$ 0.11} & 17.01 {\scriptsize $\pm$ 0.09} & 36.34 {\scriptsize $\pm$ 0.14} & 31.32 {\scriptsize $\pm$ 0.12} \\
chair  & & 38.22 {\scriptsize $\pm$ 0.21} & 21.45 {\scriptsize $\pm$ 0.16} & 44.59 {\scriptsize $\pm$ 0.25} & 53.79 {\scriptsize $\pm$ 0.27} &  & 76.97 {\scriptsize $\pm$ 0.17} & 48.26 {\scriptsize $\pm$ 0.27} & 58.58 {\scriptsize $\pm$ 0.31} & 67.79 {\scriptsize $\pm$ 0.15} &  & 34.28 {\scriptsize $\pm$ 0.18} & 17.44 {\scriptsize $\pm$ 0.13} & 33.61 {\scriptsize $\pm$ 0.16} & 42.79 {\scriptsize $\pm$ 0.19} &  & 42.30 {\scriptsize $\pm$ 0.15} & 20.63 {\scriptsize $\pm$ 0.12} & 42.38 {\scriptsize $\pm$ 0.16} & 46.83 {\scriptsize $\pm$ 0.13} \\
cabinet  & & 21.72 {\scriptsize $\pm$ 0.15} & 9.77 {\scriptsize $\pm$ 0.15} & 21.13 {\scriptsize $\pm$ 0.14} & 21.93 {\scriptsize $\pm$ 0.17} &  & 37.58 {\scriptsize $\pm$ 0.46} & 20.80 {\scriptsize $\pm$ 0.24} & 27.58 {\scriptsize $\pm$ 0.19} & 45.18 {\scriptsize $\pm$ 0.36} &  & 15.88 {\scriptsize $\pm$ 0.15} & 7.03 {\scriptsize $\pm$ 0.10} & 13.56 {\scriptsize $\pm$ 0.09} & 17.06 {\scriptsize $\pm$ 0.11} &  & 28.01 {\scriptsize $\pm$ 0.11} & 10.77 {\scriptsize $\pm$ 0.12} & 29.04 {\scriptsize $\pm$ 0.15} & 31.41 {\scriptsize $\pm$ 0.13} \\
plant  & & 18.70 {\scriptsize $\pm$ 0.09} & 15.76 {\scriptsize $\pm$ 0.20} & 10.89 {\scriptsize $\pm$ 0.13} & 47.35 {\scriptsize $\pm$ 0.44} &  & 57.32 {\scriptsize $\pm$ 0.80} & 40.09 {\scriptsize $\pm$ 0.46} & 53.77 {\scriptsize $\pm$ 0.71} & 47.22 {\scriptsize $\pm$ 0.74} &  & 16.18 {\scriptsize $\pm$ 0.13} & 12.11 {\scriptsize $\pm$ 0.08} & 9.69 {\scriptsize $\pm$ 0.10} & 30.89 {\scriptsize $\pm$ 0.46} &  & 28.54 {\scriptsize $\pm$ 0.21} & 16.47 {\scriptsize $\pm$ 0.07} & 17.65 {\scriptsize $\pm$ 0.15} & 35.55 {\scriptsize $\pm$ 0.40} \\
counter  & & 25.00 {\scriptsize $\pm$ 0.26} & 13.87 {\scriptsize $\pm$ 0.28} & 23.63 {\scriptsize $\pm$ 0.33} & 34.66 {\scriptsize $\pm$ 0.28} &  & 59.04 {\scriptsize $\pm$ 0.47} & 36.19 {\scriptsize $\pm$ 0.56} & 42.64 {\scriptsize $\pm$ 0.59} & 49.53 {\scriptsize $\pm$ 0.38} &  & 21.34 {\scriptsize $\pm$ 0.23} & 11.20 {\scriptsize $\pm$ 0.23} & 18.10 {\scriptsize $\pm$ 0.28} & 25.56 {\scriptsize $\pm$ 0.22} &  & 28.53 {\scriptsize $\pm$ 0.16} & 11.77 {\scriptsize $\pm$ 0.21} & 22.01 {\scriptsize $\pm$ 0.27} & 32.75 {\scriptsize $\pm$ 0.14} \\
sink  & & 18.85 {\scriptsize $\pm$ 0.22} & 7.78 {\scriptsize $\pm$ 0.17} & 12.52 {\scriptsize $\pm$ 0.21} & 24.64 {\scriptsize $\pm$ 0.27} &  & 44.95 {\scriptsize $\pm$ 0.64} & 19.43 {\scriptsize $\pm$ 0.23} & 55.77 {\scriptsize $\pm$ 0.41} & 41.30 {\scriptsize $\pm$ 0.60} &  & 14.84 {\scriptsize $\pm$ 0.16} & 5.65 {\scriptsize $\pm$ 0.10} & 11.35 {\scriptsize $\pm$ 0.18} & 17.71 {\scriptsize $\pm$ 0.21} &  & 31.57 {\scriptsize $\pm$ 0.15} & 10.34 {\scriptsize $\pm$ 0.14} & 22.94 {\scriptsize $\pm$ 0.19} & 30.79 {\scriptsize $\pm$ 0.21} \\

\bottomrule
\hline
\end{tabular}}
\caption{Category-level performances of \ouralg and baseline approaches in the Matterport3D test set.}
\label{tab:breakdown_results}
\end{table*}

\begin{table*}[!h]
 
\centering
\resizebox{\textwidth}{!}{
\setlength{\tabcolsep}{3pt}
\centering
\begin{tabular}{c c c c c c c c c c c c c c c c c c c c c}
\toprule
\multicolumn{2}{c}{} & \multicolumn{5}{c}{Recall} & \multicolumn{5}{c}{Precision} & \multicolumn{5}{c}{IoU} & \multicolumn{4}{c}{BF1} \\
\cmidrule(r){3-6} 
\cmidrule(r){8-11} 
\cmidrule(r){13-16} 
\cmidrule(r){18-21} 
  & & \scriptsize \shortstack{Segment $\rightarrow$ \\ Project} & \scriptsize \shortstack{Project $\rightarrow$ \\ Segment} & \scriptsize semantic SLAM &  \scriptsize \ouralg 
  & & \scriptsize \shortstack{Segment $\rightarrow$ \\ Project} & \scriptsize \shortstack{Project $\rightarrow$ \\ Segment} & \scriptsize semantic SLAM &  \scriptsize \ouralg 
  & & \scriptsize \shortstack{Segment $\rightarrow$ \\ Project} & \scriptsize \shortstack{Project $\rightarrow$ \\ Segment} & \scriptsize semantic SLAM &  \scriptsize \ouralg 
  & & \scriptsize \shortstack{Segment $\rightarrow$ \\ Project} & \scriptsize \shortstack{Project $\rightarrow$ \\ Segment} & \scriptsize semantic SLAM &  \scriptsize \ouralg  \\

\midrule
void  & & 97.77 {\scriptsize $\pm$ 0.03} & 97.34 {\scriptsize $\pm$ 0.02} & 97.58 {\scriptsize $\pm$ 0.02} & 97.00 {\scriptsize $\pm$ 0.03} &  & 91.67 {\scriptsize $\pm$ 0.04} & 85.42 {\scriptsize $\pm$ 0.07} & 91.33 {\scriptsize $\pm$ 0.04} & 93.47 {\scriptsize $\pm$ 0.04} &  & 89.79 {\scriptsize $\pm$ 0.04} & 83.46 {\scriptsize $\pm$ 0.07} & 89.30 {\scriptsize $\pm$ 0.04} & 90.84 {\scriptsize $\pm$ 0.04} &  & - & - & - \\
shelving  & & 33.11 {\scriptsize $\pm$ 0.34} & 0.00 {\scriptsize $\pm$ 0.00} & 35.08 {\scriptsize $\pm$ 0.40} & 32.24 {\scriptsize $\pm$ 0.37} &  & 62.54 {\scriptsize $\pm$ 0.49} & 0.00 {\scriptsize $\pm$ 0.00} & 59.93 {\scriptsize $\pm$ 0.64} & 69.25 {\scriptsize $\pm$ 0.99} &  & 27.31 {\scriptsize $\pm$ 0.25} & 0.00 {\scriptsize $\pm$ 0.00} & 28.05 {\scriptsize $\pm$ 0.32} & 27.60 {\scriptsize $\pm$ 0.35} &  & 27.19 {\scriptsize $\pm$ 0.25} & nan {\scriptsize $\pm$ nan} & 46.09 {\scriptsize $\pm$ 0.34} & 33.97 {\scriptsize $\pm$ 0.43} \\
dresser  & & 0.00 {\scriptsize $\pm$ 0.00} & 0.81 {\scriptsize $\pm$ 0.03} & 0.00 {\scriptsize $\pm$ 0.00} & 0.05 {\scriptsize $\pm$ 0.00} &  & 0.00 {\scriptsize $\pm$ 0.00} & 2.42 {\scriptsize $\pm$ 0.11} & 0.00 {\scriptsize $\pm$ 0.00} & 0.05 {\scriptsize $\pm$ 0.00} &  & 0.00 {\scriptsize $\pm$ 0.00} & 0.60 {\scriptsize $\pm$ 0.02} & 0.00 {\scriptsize $\pm$ 0.00} & 0.02 {\scriptsize $\pm$ 0.00} &  & nan {\scriptsize $\pm$ nan} & 2.59 {\scriptsize $\pm$ 0.06} & nan {\scriptsize $\pm$ nan} & 0.11 {\scriptsize $\pm$ 0.00} \\
bed  & & 90.55 {\scriptsize $\pm$ 0.06} & 71.00 {\scriptsize $\pm$ 0.73} & 87.22 {\scriptsize $\pm$ 0.11} & 89.39 {\scriptsize $\pm$ 0.06} &  & 86.70 {\scriptsize $\pm$ 0.45} & 55.54 {\scriptsize $\pm$ 0.75} & 79.10 {\scriptsize $\pm$ 0.56} & 85.00 {\scriptsize $\pm$ 0.47} &  & 79.43 {\scriptsize $\pm$ 0.40} & 45.46 {\scriptsize $\pm$ 0.64} & 70.81 {\scriptsize $\pm$ 0.50} & 77.14 {\scriptsize $\pm$ 0.42} &  & 65.25 {\scriptsize $\pm$ 0.32} & 32.06 {\scriptsize $\pm$ 0.39} & 69.20 {\scriptsize $\pm$ 0.24} & 61.52 {\scriptsize $\pm$ 0.40} \\
cushion  & & 61.88 {\scriptsize $\pm$ 0.25} & 11.84 {\scriptsize $\pm$ 0.23} & 44.54 {\scriptsize $\pm$ 0.29} & 62.76 {\scriptsize $\pm$ 0.30} &  & 72.68 {\scriptsize $\pm$ 0.29} & 39.80 {\scriptsize $\pm$ 0.50} & 65.86 {\scriptsize $\pm$ 0.36} & 61.51 {\scriptsize $\pm$ 0.29} &  & 50.09 {\scriptsize $\pm$ 0.23} & 9.98 {\scriptsize $\pm$ 0.19} & 35.95 {\scriptsize $\pm$ 0.23} & 44.88 {\scriptsize $\pm$ 0.22} &  & 57.75 {\scriptsize $\pm$ 0.27} & 20.09 {\scriptsize $\pm$ 0.32} & 55.88 {\scriptsize $\pm$ 0.22} & 58.55 {\scriptsize $\pm$ 0.22} \\
fireplace  & & - & - & - & - &  & - & - & - & - &  & - & - & - & - &  & - & - & - & - \\
sofa  & & 78.15 {\scriptsize $\pm$ 0.19} & 15.43 {\scriptsize $\pm$ 0.24} & 75.40 {\scriptsize $\pm$ 0.22} & 80.71 {\scriptsize $\pm$ 0.14} &  & 81.34 {\scriptsize $\pm$ 0.14} & 64.11 {\scriptsize $\pm$ 0.40} & 80.57 {\scriptsize $\pm$ 0.18} & 77.47 {\scriptsize $\pm$ 0.19} &  & 66.26 {\scriptsize $\pm$ 0.17} & 14.09 {\scriptsize $\pm$ 0.21} & 63.62 {\scriptsize $\pm$ 0.15} & 65.30 {\scriptsize $\pm$ 0.15} &  & 58.98 {\scriptsize $\pm$ 0.17} & 17.96 {\scriptsize $\pm$ 0.22} & 65.89 {\scriptsize $\pm$ 0.10} & 58.50 {\scriptsize $\pm$ 0.18} \\
table  & & 64.01 {\scriptsize $\pm$ 0.20} & 43.00 {\scriptsize $\pm$ 0.21} & 69.37 {\scriptsize $\pm$ 0.22} & 73.17 {\scriptsize $\pm$ 0.20} &  & 76.89 {\scriptsize $\pm$ 0.45} & 61.13 {\scriptsize $\pm$ 0.29} & 83.17 {\scriptsize $\pm$ 0.37} & 74.73 {\scriptsize $\pm$ 0.40} &  & 53.58 {\scriptsize $\pm$ 0.29} & 33.66 {\scriptsize $\pm$ 0.17} & 60.71 {\scriptsize $\pm$ 0.27} & 58.69 {\scriptsize $\pm$ 0.32} &  & 45.06 {\scriptsize $\pm$ 0.16} & 28.12 {\scriptsize $\pm$ 0.17} & 59.90 {\scriptsize $\pm$ 0.18} & 48.31 {\scriptsize $\pm$ 0.18} \\
chair  & & 58.77 {\scriptsize $\pm$ 0.18} & 11.00 {\scriptsize $\pm$ 0.12} & 61.79 {\scriptsize $\pm$ 0.20} & 62.31 {\scriptsize $\pm$ 0.21} &  & 69.04 {\scriptsize $\pm$ 0.24} & 29.10 {\scriptsize $\pm$ 0.39} & 77.39 {\scriptsize $\pm$ 0.20} & 76.46 {\scriptsize $\pm$ 0.17} &  & 46.38 {\scriptsize $\pm$ 0.14} & 8.67 {\scriptsize $\pm$ 0.11} & 52.28 {\scriptsize $\pm$ 0.17} & 52.21 {\scriptsize $\pm$ 0.16} &  & 53.43 {\scriptsize $\pm$ 0.17} & 12.66 {\scriptsize $\pm$ 0.13} & 62.96 {\scriptsize $\pm$ 0.21} & 58.82 {\scriptsize $\pm$ 0.16} \\
cabinet  & & 25.36 {\scriptsize $\pm$ 0.12} & 7.81 {\scriptsize $\pm$ 0.19} & 24.44 {\scriptsize $\pm$ 0.15} & 30.31 {\scriptsize $\pm$ 0.13} &  & 59.72 {\scriptsize $\pm$ 0.23} & 34.90 {\scriptsize $\pm$ 0.71} & 53.03 {\scriptsize $\pm$ 0.30} & 66.77 {\scriptsize $\pm$ 0.20} &  & 21.64 {\scriptsize $\pm$ 0.11} & 6.85 {\scriptsize $\pm$ 0.17} & 20.08 {\scriptsize $\pm$ 0.13} & 26.30 {\scriptsize $\pm$ 0.11} &  & 39.47 {\scriptsize $\pm$ 0.13} & 11.06 {\scriptsize $\pm$ 0.23} & 40.79 {\scriptsize $\pm$ 0.16} & 46.87 {\scriptsize $\pm$ 0.12} \\
plant  & & 15.83 {\scriptsize $\pm$ 0.22} & 10.43 {\scriptsize $\pm$ 0.30} & 3.56 {\scriptsize $\pm$ 0.07} & 24.99 {\scriptsize $\pm$ 0.25} &  & 99.17 {\scriptsize $\pm$ 0.02} & 70.76 {\scriptsize $\pm$ 0.72} & 97.18 {\scriptsize $\pm$ 0.14} & 98.58 {\scriptsize $\pm$ 0.02} &  & 15.80 {\scriptsize $\pm$ 0.22} & 9.95 {\scriptsize $\pm$ 0.28} & 3.56 {\scriptsize $\pm$ 0.07} & 24.90 {\scriptsize $\pm$ 0.25} &  & 29.28 {\scriptsize $\pm$ 0.28} & 11.11 {\scriptsize $\pm$ 0.32} & 6.73 {\scriptsize $\pm$ 0.14} & 33.22 {\scriptsize $\pm$ 0.29} \\
counter  & & 28.75 {\scriptsize $\pm$ 0.57} & 26.15 {\scriptsize $\pm$ 0.25} & 26.06 {\scriptsize $\pm$ 0.40} & 45.91 {\scriptsize $\pm$ 0.70} &  & 47.88 {\scriptsize $\pm$ 0.95} & 34.32 {\scriptsize $\pm$ 0.37} & 37.01 {\scriptsize $\pm$ 0.77} & 50.50 {\scriptsize $\pm$ 0.64} &  & 22.66 {\scriptsize $\pm$ 0.51} & 17.25 {\scriptsize $\pm$ 0.17} & 17.81 {\scriptsize $\pm$ 0.32} & 32.40 {\scriptsize $\pm$ 0.55} &  & 35.57 {\scriptsize $\pm$ 0.38} & 26.63 {\scriptsize $\pm$ 0.20} & 29.16 {\scriptsize $\pm$ 0.41} & 43.82 {\scriptsize $\pm$ 0.27} \\
sink  & & 23.25 {\scriptsize $\pm$ 0.55} & 26.19 {\scriptsize $\pm$ 0.52} & 24.89 {\scriptsize $\pm$ 0.34} & 40.45 {\scriptsize $\pm$ 0.81} &  & 35.13 {\scriptsize $\pm$ 0.56} & 20.98 {\scriptsize $\pm$ 0.46} & 22.04 {\scriptsize $\pm$ 0.67} & 22.58 {\scriptsize $\pm$ 0.36} &  & 16.01 {\scriptsize $\pm$ 0.37} & 12.06 {\scriptsize $\pm$ 0.23} & 12.46 {\scriptsize $\pm$ 0.29} & 15.54 {\scriptsize $\pm$ 0.21} &  & 46.68 {\scriptsize $\pm$ 0.70} & 22.49 {\scriptsize $\pm$ 0.26} & 28.25 {\scriptsize $\pm$ 0.41} & 34.77 {\scriptsize $\pm$ 0.37} \\

\bottomrule
\hline
\end{tabular}}
\caption{Category-level performances of \ouralg and baseline approaches in the Replica dataset. Note that the \textit{fireplace} category is not present in the Replica dataset.}
\label{tab:breakdown_results_replica}
\end{table*}


\subsection{Navigation path recording interface}
\begin{figure*}[h]
    \centering
    \caption{Example navigation trajectory and collection interface.} 
    \includegraphics[width=1.0\linewidth, clip]{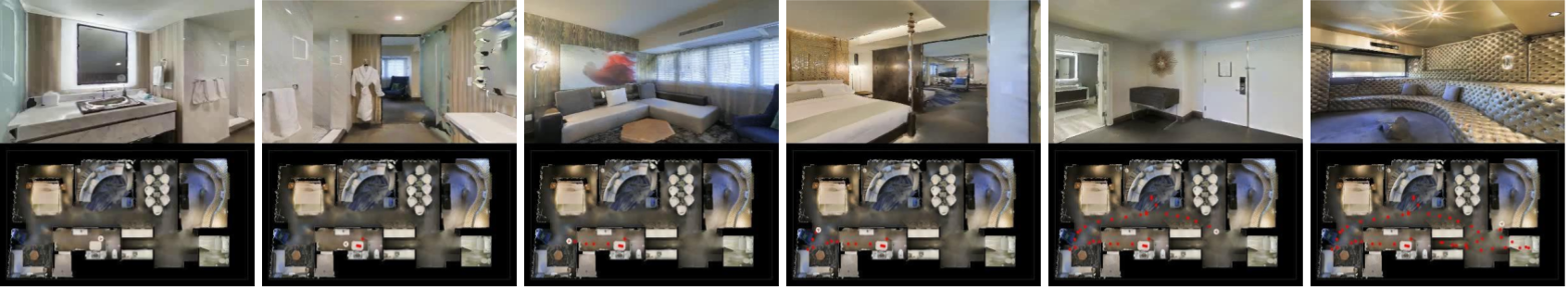}
\label{fig:navpath}
\end{figure*}

Figure \ref{fig:navpath} represents an example navigation trajectory. We manually recorded these paths by leveraging the top-down map during navigation collection to encourage trajectories that provide a good coverage of the scene.

\subsection{\ouralg visual feature encoder. Experimentation of using different egocentric features extracted at different stages in RedNet.}
In this section we study the impact of projecting egocentric features extracted at different stages in RedNet \cite{jiang2018rednet}. Specifically we extract the last encoder layer features, the last layer of the decoder features, the scores, the probability distribution over the 13 categories (softmax of the scores) and the one-hot encoded version of the egocentric labels (we one-hot encoded the argmax of the scores).
In addition, we also vary the number of channels in the memory tensor. For each feature type, we experiment with the following two settings (1) the number of memory channels is equal to the number of egocentric feature channels. (2) The number of memory channels equals to 256. We set that number to be larger than the number of channels of the projected features (expect for the encoder features) in order to allow more space for the GRU to accumulate information over time.
\tableref{tab:smnet_ablation_diff_features} groups all of these experiments with boostrapped standard error. We also report the egocentric resolution of the different features. For all these experiment we up-sample the egocentric feature map resolution to (480 $\times$ 640) using bi-linear interpolation before projection (if necessary).
Note that the two experiments of projecting the last encoder layer features with 256 and 512 number of memory channels are missing due to memory constraint during training.

We find that projecting the last layer features of the RedNet decoder \cite{jiang2018rednet} with 256 number of memory channels yields to the best performances (mIoU 36.49, mBF1 36.70).

\begin{table*}[!h]
\centering
\resizebox{\textwidth}{!}{
\begin{tabular}{ c c c c c c c c c c s s }
\toprule
  & & \shortstack{What's \\ Projected} & \shortstack{Ego. feat. \\ resolution} & \shortstack{Features \\ \#Channels}  & \shortstack{Memory \\ \#Channels} & & Acc & mRecall & mPrecision & mIoU & mBF1 \\ 
  \midrule
   \multirow{9}{*}{\rotatebox{90}{\ouralg}} & & 
   encoder & (15$\times$20) & 512 & 128 & & 85.82 {\scriptsize $\pm$ 0.06} & 38.97 {\scriptsize $\pm$ 0.06} & 52.66 {\scriptsize $\pm$ 0.09} & 29.03 {\scriptsize $\pm$ 0.06} & 26.24 {\scriptsize $\pm$ 0.05} \\
& & last layer & (240$\times$320) & 64 & 64 &  & 87.92 {\scriptsize $\pm$ 0.08} & 46.89 {\scriptsize $\pm$ 0.11} & 58.46 {\scriptsize $\pm$ 0.10} & 36.27 {\scriptsize $\pm$ 0.08} & 36.35 {\scriptsize $\pm$ 0.09} \\
& & last layer & (240$\times$320) & 64 & 256  & & \textbf{88.18 {\scriptsize $\pm$ 0.09}} & \textbf{47.20 {\scriptsize $\pm$ 0.11}} & 57.99 {\scriptsize $\pm$ 0.11} & \textbf{36.49 {\scriptsize $\pm$ 0.09}} & \textbf{36.70 {\scriptsize $\pm$ 0.09}} \\
& & scores & (480$\times$640) & 13   & 13  &  & 87.94 {\scriptsize $\pm$ 0.08} & 41.25 {\scriptsize $\pm$ 0.10} & 59.34 {\scriptsize $\pm$ 0.09} & 33.42 {\scriptsize $\pm$ 0.08} & 33.78 {\scriptsize $\pm$ 0.08} \\
& & scores & (480$\times$640)  & 13  & 256  &  & 87.83 {\scriptsize $\pm$ 0.09} & 43.10 {\scriptsize $\pm$ 0.11} & \textbf{59.46 {\scriptsize $\pm$ 0.11}} & 34.63 {\scriptsize $\pm$ 0.09} & 35.75 {\scriptsize $\pm$ 0.09} \\
& & softmax & (480$\times$640)  & 13  & 13  &  & 87.86 {\scriptsize $\pm$ 0.06} & 41.76 {\scriptsize $\pm$ 0.08} & 58.70 {\scriptsize $\pm$ 0.13} & 33.47 {\scriptsize $\pm$ 0.06} & 33.15 {\scriptsize $\pm$ 0.07} \\
& & softmax & (480$\times$640)  & 13  & 256  &  & 87.79 {\scriptsize $\pm$ 0.07} & 43.25 {\scriptsize $\pm$ 0.09} & 59.41 {\scriptsize $\pm$ 0.11} & 34.38 {\scriptsize $\pm$ 0.07} & 34.30 {\scriptsize $\pm$ 0.08} \\
& & one-hot & (480$\times$640)  & 13  & 13  &  & 87.85 {\scriptsize $\pm$ 0.06} & 40.68 {\scriptsize $\pm$ 0.08} & 58.01 {\scriptsize $\pm$ 0.13} & 32.52 {\scriptsize $\pm$ 0.06} & 32.64 {\scriptsize $\pm$ 0.07} \\
& & one-hot & (480$\times$640)  & 13  & 256  &  & 87.52 {\scriptsize $\pm$ 0.07} & 44.65 {\scriptsize $\pm$ 0.08} & 56.07 {\scriptsize $\pm$ 0.12} & 34.12 {\scriptsize $\pm$ 0.07} & 33.47 {\scriptsize $\pm$ 0.07}  \\
\bottomrule
 \end{tabular}}
  \caption{Results of \ouralg on top-down semantic segmentation on the Matterport3D dataset under different settings. Here we experiment with different egocentric features extracted at different stages in RedNet. We also vary the number of channels in the memory tensor.}
 \label{tab:smnet_ablation_diff_features}
 \end{table*}

\subsection{Heuristic to reduce ``label splatter'' in  Segment $\rightarrow$ Project}
In this section we explore simple processing techniques to enhance the performances of the Seg. $\rightarrow$ Proj. baseline. 
In the Matterport3D dataset \cite{chang2017matterport3d} the depth values on the edges of objects can sometimes be unreliable. This is due to slightly different geometry between the depth and semantic meshes. As a result, in the Segment $\rightarrow$ Project approach, the projected segmentation maps can be noisy and splatter around the map. 
In addition and in general, any egocentric labeling mistakes made around objects boundaries will lead to the same error. 
To reduce the impact of this problem, we try to (1) erode the semantic prediction map with a 10 pixel square filter. We apply a binary erosion filter to the semantic prediction frame for all categories. (2) Downsample the egocentric semantic prediction map from (480 $\times$ 640) to (120 $\times$ 160). Any missed pixels in the observed area of the top-down semantic map caused by this down-sampling is filled using median filtering (k=10). (3) Post-process the top-down semantic maps with a median filter (k=3).
However, note that those are simple heuristics and may not generalize well to new data collected by other sensors.

We find that applying erosion on the egocentric semantic predicton map hurts the performances: mIoU 29.34 $>$ 23.07 at full resolution (480$\times$640) and mIoU 31.25 $>$ 24.29 at low resolution (120$\times$160). However, down-sampling the semantic frame boosts the performances: mIoU 31.25 $>$ 29.34. In addition, post-processing the semantic maps with a median filter improves even further the results on the mIoU metric 32.49 $>$ 31.25.  

In our experiments we find that \ouralg performs better than the Seg. $\rightarrow$ Proj. baseline – achieving mBF1 37.02, and mIoU 36.77. We find that applying a decoder on top of the memory tensor helps reducing the label splatter phenomena.


\begin{table*}[!h]
\centering
\resizebox{\textwidth}{!}{
\begin{tabular}{ c c c c c c c c c s s }
\toprule
  & & \shortstack{Egocentric \\ resolution} & \shortstack{Egocentric \\ erosion} & \shortstack{post-processing \\ med. filter}  & & Acc & mRecall & mPrecision & mIoU & mBF1 \\ 
  \midrule
   \multirow{5}{*}{\rotatebox{90}{Seg. $\rightarrow$ Proj.}} & & 
    (480$\times$640) & - & - & & 86.08 {\scriptsize $\pm$ 0.05} & 37.47 {\scriptsize $\pm$ 0.07} & 51.86 {\scriptsize $\pm$ 0.09} & 29.34 {\scriptsize $\pm$ 0.05} & 33.08 {\scriptsize $\pm$ 0.06} \\
& & (480$\times$640) & $\checkmark$ & - &  & 85.35 {\scriptsize $\pm$ 0.05} & 26.70 {\scriptsize $\pm$ 0.06} & 57.59 {\scriptsize $\pm$ 0.11} & 23.07 {\scriptsize $\pm$ 0.05} & 27.95 {\scriptsize $\pm$ 0.06} \\

& & (120$\times$160) & $\checkmark$ & - &  & 85.81 {\scriptsize $\pm$ 0.05} & 28.00 {\scriptsize $\pm$ 0.06} & \textbf{59.49 {\scriptsize $\pm$ 0.12}} & 24.29 {\scriptsize $\pm$ 0.05} & 28.11 {\scriptsize $\pm$ 0.06}\\
& & (120$\times$160) & - & - &  & 87.07 {\scriptsize $\pm$ 0.06} & 39.48 {\scriptsize $\pm$ 0.07} & 54.62 {\scriptsize $\pm$ 0.10} & 31.25 {\scriptsize $\pm$ 0.06} & \textbf{34.22 {\scriptsize $\pm$ 0.07}}\\
& & (120$\times$160) & - & $\checkmark$ &  & \textbf{88.14 {\scriptsize $\pm$ 0.07}} & \textbf{40.18 {\scriptsize $\pm$ 0.09}} & 58.70 {\scriptsize $\pm$ 0.11} & \textbf{32.49 {\scriptsize $\pm$ 0.07}} & 33.03 {\scriptsize $\pm$ 0.07}\\
\bottomrule
 \end{tabular}}
  \caption{Results of the Seg. $\rightarrow$ Proj. baseline on top-down semantic segmentation on the Matterport3D dataset under different settings.}
 \label{tab:segproj_results}
 \end{table*}

\subsection{Object Navigation in more detail.}
For the ObjectNav downstream task we opt for an open loop planing strategy using A*. 
First we compute a free space map of the environment using the map of heights. The map of heights is generated while running \ouralg. At each time step, the height of each projection is stored in this map. Recall that for pixels in a given frame projecting onto the same 2D location in the top-down map we keep the pixel with highest height. In addition, points above 50cm of the camera height are discarded to avoid detecting ceiling pixels.
Using the map of heights we estimate the floor height as the predominant height value. We then threshold the map of heights around the floor height ($\pm$ 5cm) to create a free space map. We preprocess the free space map using a binary closing operation with a square element of 10 pixels in order to create a contiguous map.

Secondly, we create a binary goal map from the predicted top-down semantic map by setting to 1 any pixels labeled as the target object category. We preprocess the goal map with a binary opening operation with a square element of 10 pixels to limit the impact of small noisy detections in the top-down semantic map.
We then use A* with the free space map and the goal map to find a path from the start location to the nearest target object instance. Any locations equal to 1 in the free space map are potential nodes in A*. At each time step our planner looks at all neighbor nodes and select the one closer to target based on a Euclidean metric. The neighbors' locations are defined on a 0.25m radius circle centered on the agent's position and spawn equally every 30deg from the agent's current heading. This accounts for the set of actions with step sizes defined in the Habitat challenge \cite{habitat_challenge20} (i.e forward step size of 0.25m and rotation angle of 30deg). We filter out unreachable neighbors by testing if the agent can navigate from its position to the neighbor's position. We perform this by dragging a disk of radius equal to the agent's radius along the line between the agent's position and the neighbor's position on the free space map.
Lastly, a target object is reached if the current node is within a 1m radius and the object can be seen by the agent. In order to implement the second part of the stopping criteria we select, from the goal map, each target pixels within a 1m range from the current node. The goal is considered reached if for one of those selections, the pixels along the line between the agent and the given goal pixel are all observed (an observed pixel is a pixel that has received a projection).

Ultimately, we run the generated trajectories in the Habitat simulator (Savva et al.
2019) for evaluation.
We evaluate this strategy on the validation set of the ObjectNav Habitat challenge \cite{habitat_challenge20}.
Using this approach we achieve a success rate of $9.658 \%$, with SPL of $5.714 \%$, soft SPL of $ 8.702\%$ and average distance to the target of $7.31576m$. Note that $26 \%$ of the episodes in this set are targeting object categories falling outside of our list of object classes, we consider those as failure. The evaluation metrics limited to episodes targeting objects in our list of classes are: success rate of $13.070 \%$, with SPL of $7.733 \%$, soft SPL of $11.777\%$ and average distance to the target of $6.70981m$. \tableref{tab:objnav_breakdown} breakdowns the metrics per house. There are 5 houses (\texttt{8194nk5LbLH, oLBMNvg9in8, QUCTc6BB5sX, X7HyMhZNoso, x8F5xyUWy9e}) with a success rate lower than 0.05 while the rest are greater than 0.10. Part of the explanation is this is due to naive estimations of free space for those 5 houses.

\begin{table}[!h]
\centering
\resizebox{\linewidth}{!}{
\begin{tabular}{ c c c c c c c c c}
\toprule
 & \shortstack{GT free \\ space}  & & $\#$ episode & dist. to goal & success  & SPL & soft SPL\\
\midrule
\multirow{2}{*}{ALL episodes} & - & & 2195 & 7.31576 & 0.09658 & 0.05714 & 0.08702 \\
                                              & $\checkmark$ & & 2195 & 5.65880 & 0.31250 & 0.20700 & 0.28290\\
\midrule
\multirow{2}{*}{\shortstack{Episodes w/ \\ \ouralg Obj. Cat.}}& -& & 1622 & 6.70981 & 0.1307 & 0.07733 & 0.11777 \\
                                              & $\checkmark$ & & 1622 & 4.46750 & 0.42290 & 0.28010 & 0.38280\\
\bottomrule
 \end{tabular}}
  \caption{ObjectNav results comparison with an A* planner using the ground truth free space maps.}
 \label{tab:objnav_GTfreespace}
 \end{table}

In order to estimate the impact of the free space map estimation on the results, we run our planner with the ground truth free space maps. The ground truth free space maps are generated using the Habitat API \cite{habitat19iccv}.
\tableref{tab:objnav_GTfreespace} shows results of this ablation experiment. Using the ground-truth free space maps we see an increase in success of +0.21592 and SPL of +0.14986 on all episodes and an increase in success of +0.2922 and SPL of +0.20277 on episodes targeting objects in our list of classes. \tableref{tab:objnav_breakdown_GTfreespace} shows results per house with ground-truth free space maps. On the previous 5 outlier houses we see an average improvement on success of +0.24244 and SPL of +0.17444.

We now study the impact of the predicted top-down maps on the performances. We select 7 out of the 22 validation environments (\texttt{zsNo4HB9uLZ\_0, x8F5xyUWy9e\_0, X7HyMhZNoso\_0, QUCTc6BB5sX\_0, QUCTc6BB5sX\_1, 8194nk5LbLH\_0, Z6MFQCViBuw\_0}) for which we have ground-truth top-down semantic maps (the remaining 15 environments are discarded because they belong to houses hard to divide by floor using a simple plane). We run our algorithm for those 7 environments with the following information provided to the planner: 
(1) ground-truth free space and ground-truth semantic maps, 
(2) ground-truth free space and predicted semantic maps, 
(3) predicted free space and ground-truth semantic maps and, 
(4)predicted free space and predicted semantic maps. \tableref{tab:objnav_GTfreespace_GTsemmaps} groups all results.

 \begin{table}[!h]
\centering
\resizebox{\linewidth}{!}{
\begin{tabular}{ c c c c c c c c c c }
\toprule
  & & \shortstack{GT semantic \\ maps} &  \shortstack{GT free \\ space }  &  & dist. to goal & success  & SPL & soft SPL\\ 
 \midrule
 \multirow{5}{*}{\shortstack{771 Episodes  \\ (7 Env.)}}& (1) & $\checkmark$ & $\checkmark$ & & 1.8805 & 0.6913 & 0.4993 & 0.6154\\
                                                      & (2) & - & $\checkmark$ & & 4.1706 & 0.4306 & 0.2924 & 0.4177\\
                                                      & (3) & $\checkmark$ & - & & 6.4279 & 0.2153 & 0.1403 & 0.1777\\
                                                      & (4) & - & - & & 7.3593 & 0.1154 & 0.0669 & 0.0967\\
\bottomrule
 \end{tabular}}
  \caption{ObjectNav results on a subset of episodes (771) from the validation set of the ObjectNav Habitat challenge \cite{habitat_challenge20}. This table compares performances of our A* planner when the ground truth free space maps and the ground-truth semantic maps are provided.}
 \label{tab:objnav_GTfreespace_GTsemmaps}
 \end{table}

On this subset of episodes, our algorithm achieves a success of 0.6913 when the ground-truth free space and semantic maps are provided. On a rough estimate, from the 30$\%$ failure cases, 10$\%$ are due to the planner and the other 20$\%$ are due to failures in the simulator. We find that the planner failure cases are mostly due to very narrow sections on the free space map making it hard for the algorithm to plan a path through considering some radius margin for the agent. In simulation, the agent very rarely does collisions. However, the major source of error is due to a delicate implementation of episode success in A* leading to wrong final agent's position.
When comparing experiments (1) and (2) form \tableref{tab:objnav_GTfreespace_GTsemmaps} we see a decrease in performances of -0.2607 in success and -0.2069 in SPL caused by prediction errors on the top-down semantic map. When comparing (1) and (3) we witness a much larger drop in performances with -0.476 in success and -0.359 SPL caused by errors on the free space maps. Experiment (4) reports results of using both the predicted semantic maps and free space maps. Errors on those two predicted maps reduce the performances of -0.5759 on success and -0.4324 on SPL compared to (1).

These results suggest that the estimation of the free space is the major source of error. This can be improved by extended \ouralg to predict free space. In addition, these outcomes suggest that the predicted top-down semantic maps contains useful spatial and semantic information and allow good performances on the ObjecNav task \cite{habitat_challenge20} in this pre-exploration setting.

\begin{table*}[!h]
\centering
\resizebox{\textwidth}{!}{
\begin{tabular}{ c c c c c c c c c c c c c c }
\toprule
 & ALL episodes& &2azQ1b91cZZ & 8194nk5LbLH & EU6Fwq7SyZv & oLBMNvg9in8 & pLe4wQe7qrG & QUCTc6BB5sX & TbHJrupSAjP & X7HyMhZNoso & x8F5xyUWy9e & Z6MFQCViBuw & zsNo4HB9uLZ\\
 \midrule
 $\#$ episodes & 2195 & & 200 & 201 & 198 & 201 & 200 & 197 & 200 & 200 & 200 & 200 & 198\\
 distance to goal & 7.31576 & & 6.3678 & 13.57 & 6.2462 & 4.9034 & 2.1607 & 11.4774 & 7.8346 & 11.0309 & 6.4799 & 6.5443 & 3.8561\\
 success & 0.09658 & & 0.205 & 0.0398 & 0.101 & 0.0199 & 0.105 & 0.0457 & 0.145 & 0.04 & 0 & 0.11 & 0.2525\\
 SPL & 0.05714 & & 0.1168 & 0.0249 & 0.0479 & 0.0113 & 0.0579 & 0.0127 & 0.1083 & 0.0275 & 0 & 0.0661 & 0.1559\\
 soft SPL & 0.08702 & & 0.2502 & 0.0352 & 0.0706 & 0.0221 & 0.0578 & 0.0389 & 0.1371 & 0.0447 & 0.0001 & 0.1161 & 0.1851\\
\bottomrule
 \end{tabular}}
  \caption{Per house breakdown results on the validation set of the ObjectNav Habitat challenge \cite{habitat_challenge20}.}
 \label{tab:objnav_breakdown}
 \end{table*}

\begin{table*}[!h]
\centering
\resizebox{\textwidth}{!}{
\begin{tabular}{ c c c c c c c c c c c c c c }
\toprule
 & ALL episodes& &2azQ1b91cZZ & 8194nk5LbLH & EU6Fwq7SyZv & oLBMNvg9in8 & pLe4wQe7qrG & QUCTc6BB5sX & TbHJrupSAjP & X7HyMhZNoso & x8F5xyUWy9e & Z6MFQCViBuw & zsNo4HB9uLZ\\
 \midrule
 $\#$ episodes & 2195 & & 200 & 201 & 198 & 201 & 200 & 197 & 200 & 200 & 200 & 200 & 198\\
 distance to goal & 5.6588 & & 4.5287 & 12.9276 & 5.4649 & 4.1525 & 1.6194 & 7.7236 & 6.4149 & 7.4915 & 3.5627 & 5.4019 & 2.9319\\
 success & 0.3125 & & 0.37 & 0.0547 & 0.2879 & 0.1791 & 0.5 & 0.2234 & 0.33 & 0.375 & 0.38 & 0.34 & 0.399 \\
 SPL & 0.207 & & 0.231 & 0.0369 & 0.1655 & 0.1111 & 0.3269 & 0.166 & 0.2342 & 0.2998 & 0.2584 & 0.2029 & 0.2449\\
 soft SPL & 0.2829 & & 0.3838 & 0.0621 & 0.249 & 0.1592 & 0.3204 & 0.3517 & 0.2912 & 0.4052 & 0.3234 & 0.2315 & 0.3374\\
\bottomrule
 \end{tabular}}
  \caption{Per house breakdown results on the validation set of the ObjectNav Habitat challenge \cite{habitat_challenge20} with ground-truth free space maps.}
 \label{tab:objnav_breakdown_GTfreespace}
 \end{table*}

